\documentclass[acmtog]{acmart}

\setcitestyle{square}

\usepackage{colortbl}

\usepackage{times}
\usepackage{epsfig}
\usepackage{graphicx}
\usepackage{amsmath}
\usepackage{amssymb}

\usepackage{subfig}
\usepackage{mathtools}
\usepackage{amsfonts}
\usepackage{float}

\usepackage{filecontents}

\usepackage[T1]{fontenc}
\usepackage[utf8]{inputenc}

\usepackage{graphicx}
\usepackage{tikz}
\usetikzlibrary{matrix}

\usepackage{listings}

\usepackage{caption}
\usepackage{hyperref}

\newcommand{\etal}{\textit{et al. }}

\tikzset{square matrix/.style={
    matrix of nodes,
    column sep=-\pgflinewidth, row sep=-\pgflinewidth,
    nodes={draw,
      minimum height=#1,
      anchor=center,
      text width=#1,
      align=center,
      inner sep=0pt
    },
  },
  square matrix/.default=5.8mm
}

\newcommand{\financer}{Adidas AG\textsuperscript{TM}}
\begin{document}

\title{A complete hand-drawn sketch vectorization framework}

%

\author{Luca Donati}
\affiliation{%
  \institution{University of Parma}
  \streetaddress{Parco Area delle Scienze, 181/A}
  \city{Parma}
  \postcode{43124}
  \country{Italy}}
\email{luca.donati@unipr.it}
\author{Simone Cesano}
\affiliation{%
  \institution{Adidas AG}
  \streetaddress{Adi-Dassler-Strasse 1}
  \city{Herzogenaurach}
  \postcode{91074}
  \country{Germany}}
\email{simone.cesano@adidas.com}
\author{Andrea Prati}
\orcid{0000-0002-1211-529X}
\affiliation{%
  \institution{University of Parma}
  \streetaddress{Parco Area delle Scienze, 181/A}
  \city{Parma}
  \postcode{43124}
  \country{Italy}}
\email{andrea.prati@unipr.it}

\renewcommand\shortauthors{Donati, L. et al}

\begin{abstract}
Vectorizing hand-drawn sketches is a challenging task, which is of paramount importance for creating CAD vectorized versions for the fashion and creative workflows. This paper proposes a complete framework that automatically transforms noisy and complex hand-drawn sketches with different stroke types in a precise, reliable and highly-simplified vectorized model.

The proposed framework includes a novel line extraction algorithm based on a multi-resolution application of Pearson's cross correlation and a new unbiased thinning algorithm that can get rid of scribbles and variable-width strokes to obtain clean 1-pixel lines. Other contributions include variants of pruning, merging and edge linking procedures to post-process the obtained paths. Finally, a modification of the original Schneider's vectorization algorithm is designed to obtain fewer control points in the resulting Bezier splines. 

All the proposed steps of the framework have been extensively tested and compared with state-of-the-art algorithms, showing (both qualitatively and quantitatively) its outperformance.
\end{abstract}


\makeatletter

\makeatother
\maketitle

\section{Introduction}
Raw paper sketches are usually the starting point of many creative and fashion workflows. For many artists the choice of drawing with pencil and paper (or pen) grants them the most expressiveness and creative freedom possible. By using these simple tools they can convey ideas in a very fast and natural way. That allows them to propose powerful and innovative designs. Later, the prototypal idea from the hand-drawn sketch must be converted to a real world product.

The \textit{de-facto} standard for distributing fashion and mechanical designs is the vectorized set of lines composing the raw sketches: formats like SVG, CAD, Adobe Illustrator files are manually created by the designers, and delivered and used by a plethora of departments for many applications, e.g. marketing, production line, end-of-season analyses, etc..

Unfortunately, the vectorization process is still a manual task, which is both tedious and time-consuming. Designers have to click over each line point by point and gain a certain degree of experience with the used tools to create a good representation of the original sketch model.

Therefore, the need for an automated tool arises. A great amount of designer's time can be saved and re-routed in the creative part of their job. Some tools with this specific purpose are commercially available, such as Adobe Illustrator\textsuperscript{TM}\footnote{\url{https://www.adobe.com/products/illustrator.html}} Live Trace, Wintopo\footnote{\url{http://wintopo.com/}} and Potrace\footnote{\url{http://potrace.sourceforge.net/}}. Anyway, to our experience and knowledge with Adidas designers, none of these tools does this job in a proper or satisfying way.

Vectorization of hand-drawn sketches is a well researched area, with robust algorithms, such as SPV \cite{dori1999sparse} and OOPSV \cite{song2002object}. However, these methods, as well as others in the literature, fail to work with real scribbles composed of multiple strokes, since they tend to vectorize each single line, while not getting the right semantics of the drawing \cite{bartolo2007scribbles}. Problems to be faced in real cases are mainly the followings: bad/imprecise line position extractions; lines merged together when they should not or split when they were a single one in the sketch; lines extracted as ``large'' blobs (shapes with their relative widths instead of zero-width line segments); unreliability of detection with varying stroke hardness (dark lines are overly detected, faint lines are not detected at all); resulting ``heavy'' b-splines (the vectorized shapes are composed of too many control points, making subsequent handling hard).

Some works for a complete vectorization workflow are also present in the literature, even if mainly addressing vectorization of really clean sketches, or obtaining decent artistic results from highly noisy and ``sloppy'' paintings. None of them is designed to work with ``hard'' real data trying to retrieve the most precise information of exact lines and produce high quality results.

This paper provides a complete workflow for automated vectorization of hand-drawn sketches. Two new methods are presented: a reliable line extraction algorithm and a fast unbiased thinning. Moreover, many different existing techniques are discussed, improved and evaluated: paths extraction, pruning, edge linking, and Bezier curve approximation.

The efficacy of the proposal has been demonstrated on both hand-drawn sketches and images with added artificial noise, showing in both cases excellent performance w.r.t. the state of the art. Feedbacks from Adidas designers testify the quality of the process and prove that it greatly outperforms existing solutions, in terms of both efficiency and accuracy.

The remainder of this paper is structured as follows. The next Section reports the related work in sketch vectorization. Section \ref{sec:framework} describes the different steps of the framework (namely, line extraction, thinning, path creation and vectorization). Experimental results on these steps are reported in Section \ref{sec:experiments}, while Section \ref{sec:conclusions} summarizes the contributions and draws some conclusions.

\section{Related work}
This section reports the most relevant previous works on sketch vectorization. \cite{bartolo2007scribbles} proposed a line enhancement method, based on Gabor and Kalman filters. It can be used to enhance lines for subsequent vectorization. However, this approach fails to correctly extract all the drawing components when the image is noisy or presents parallel strokes, resulting, for instance, in gaps in the final vectorized result or strokes incorrectly merged. Moreover, experiments are conducted with quite simple images.

\cite{hilaire2006robust} reported a first proposal of a framework transforming raw images to full vectorized representations. However, the ``binarization'' step is not considered at all, by presenting directly the skeleton processing and vectorization steps. In addition to this limitation, this paper also bases the vectorization to the simple fitting of straight lines and circular arcs (instead of using Bezier interpolation), which represents a too simplified and limited representation of the resulting path.

\cite{noris2013topology} provided a more complete study of the whole vectorization process. They provide a neat derivation-based algorithm to estimate accurate centerlines for sketches. They also provide a good insight of the problem of correct junction selection. Unfortunately, they work under the assumption of somewhat ``clean'' lines, that does not hold in many real case scenarios, such as those we are aiming at.

\cite{SimoSerraSIGGRAPH2016} trained a Convolutional Neural Network to automatically learn how to simplify a raw sketch. No preprocessing or postprocessing is necessary and the network does the whole process of conversion from the original sketch image to a highly-simplified version. This task is related to our needs, since it can be viewed as a full preliminary step, that just needs the vectorization step to provide the final output form. In Section \ref{line_ex_results} we will compare with this work and show that our proposal achieves better results for retrieval (in particular in terms of recall) and is more reliable when working with different datasets and input types.

\cite{favreau2016fidelity} provided an overview of the whole process. Anyway, they gave just brief notions of line extraction and thinning, while they concentrated more on the final vectorization part, in which they proposed an interesting global fitting algorithm. Indeed, this paper is the only paper providing guidelines to obtain for the final output a full Bezier-curves representation. Representing images with Bezier curves is of paramount importance in our application domain (fashion design), and is moreover important to obtain ``lightweight'' vectorial representations of the underlying shapes (composed of as few control points as possible). The vectorization task is casted as a graph optimization problem. However, they treated just partially the noisy image problem, focusing on working with somewhat cleaner paintings.

Another recent work, \cite{bessmeltsev2018vectorization}, provided a good proposal for a vectorization system. For the line extraction part they rely on Vector Fields, which give high quality results with clean images, but fail in presence of noise and fuzzy lines. Still, they dedicated a lot of attention to correctly disambiguate junctions and parallel strokes.

The sketch vectorization field also partially overlaps with the so-called ``Coherence Enhancing'' field. \cite{kang2007coherent} estimated Tangent Vector Fields from images, and used them in order to clean or simplify the input. They do that by averaging a pixel value with its corresponding neighbors along the Vector Fields. This could be integrated as a useful preprocessing step in our system, or could be used as a standalone tool if the objective is just to obtain a simplified representation of the input image.

The same research topic has been explored in the work from \cite{chen2013non}, that showed remarkable results in the task of disambiguating parallel, almost touching, strokes (a very useful property in the sketch analysis domain).

\section{Sketch vectorization Framework}\label{sec:framework}
We propose a modular workflow for sketch vectorization, differently from other papers (e.g., \cite{SimoSerraSIGGRAPH2016}) that are based on monolithic approaches. Following a monolithic approach usually gives very good results for the specific-task dataset, but grants far less flexibility and adaptability, failing to generalize for new datasets/scenarios. A modular workflow is better from many perspectives: it is easy to add parts to the system, change them or adapt them to new techniques and it is much easier to maintain their implementations, and to parametrize or add options to them (expected error, simplification strength, etc..).

\begin{figure}
  \centering
  \subfloat[][]{
  \includegraphics[width=0.5\textwidth]{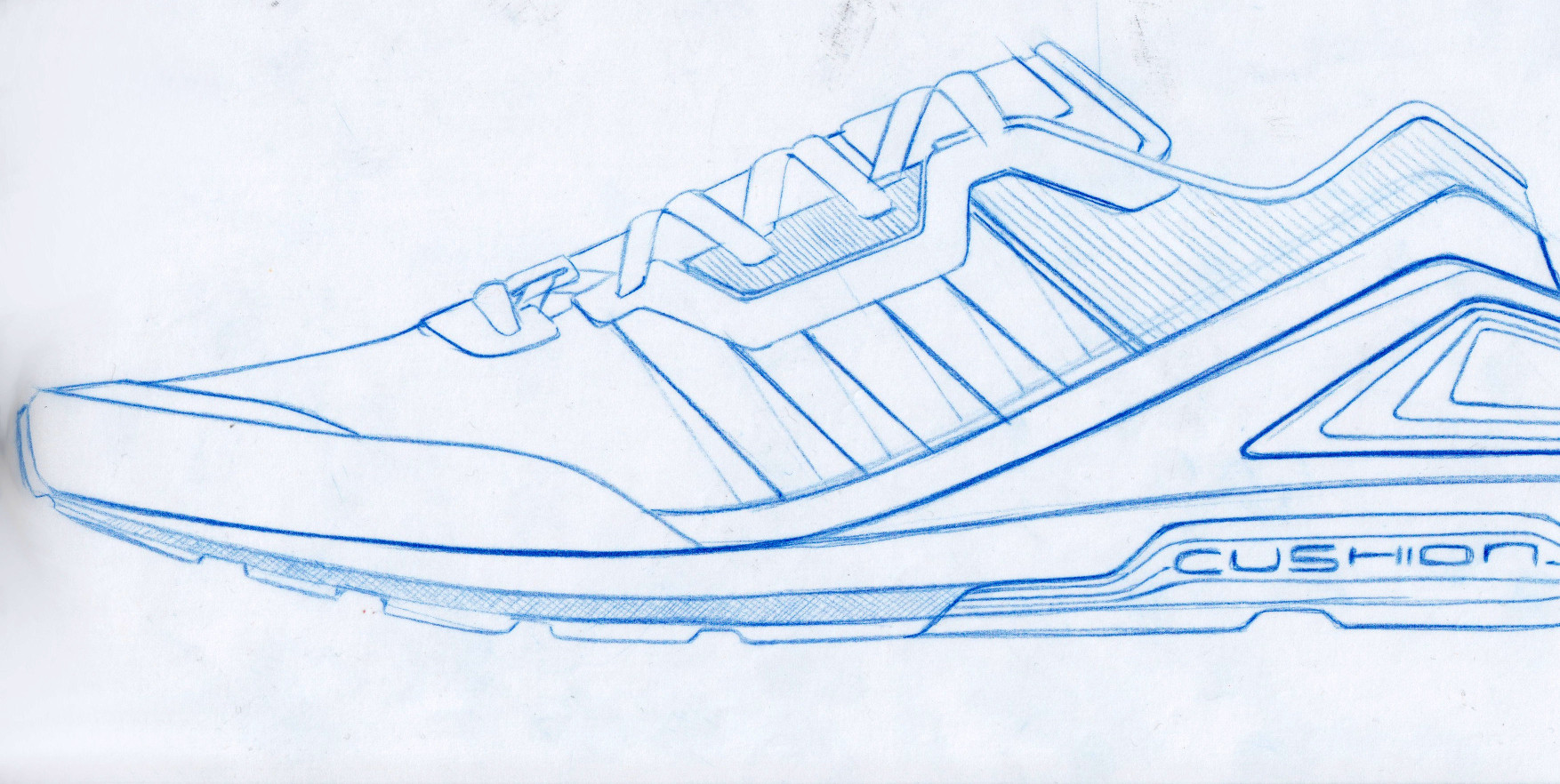} \label{good}
  }
  
  \subfloat[][]{
  \includegraphics[width=0.5\textwidth]{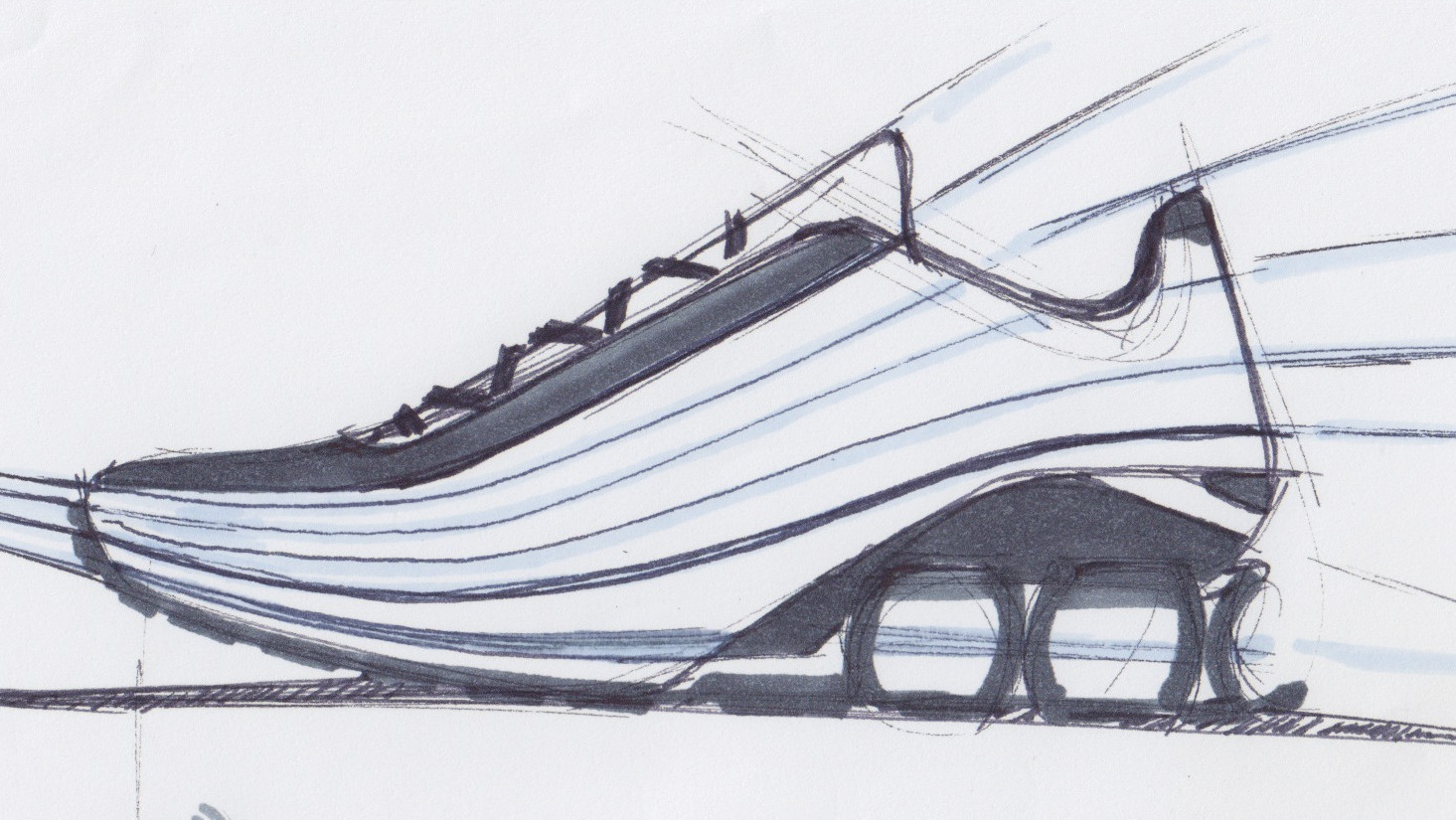} \label{bad}
  }
  \caption{Examples of the kind of sketches we treat \protect\subref{good} (monochrome, lines only), and another type that is not subject of this article \protect\subref{bad} (color, large dark blobs).}
  \label{fig:goodbadshoe}
\end{figure}

The approach proposed in this paper starts from the assumption that the sketch is a monochromatic, lines-only image. That is, we assume that no ``dark'' large blob is present in the area, just traits and lines to be vectorized. Fig. \ref{good} shows an example of the sketches we aim to vectorize, while Fig. \ref{bad} reports another exemplar sketch that, given the previous assumption, will not be considered in this paper. Moreover, we will also assume a maximum allowed line width to be present in the input image. These assumptions are valid in the vast majority of sketches that would be useful vectorize: fashion designs, mechanical parts designs, cartoon pencil sketches and more.

Our final objective is to obtain a vectorized representation composed by zero-width curvilinear segments (Bezier splines or b-splines), with as few control points as possible.

This is an overview of the modules composing the workflow:
\begin{itemize}
\item First, line presence and locations from noisy data, such as sketch paintings, need to be extracted. Each pixel of the input image will be labeled as either part of a line or background.
\item Second, these line shapes are transformed into 2d paths (each path being an array of points). This can be done via a thinning and some subsequent post-processing steps.
\item Third, these paths are used as input data to obtain the final vectorized b-splines.
\end{itemize}

Each of these modules will be extensively described in the next subsections.

\subsection{Line extraction}
Extracting precise line locations is the mandatory starting point for the whole vectorization process. When working with hand-drawn sketches, we usually deal with pencil lines traced over rough paper. Other options are pens, marker pens, ink or PC drawing tables.

\begin{figure}
  \centering
  \includegraphics[width=0.6\columnwidth]{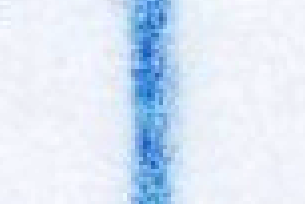}
  \includegraphics[width=0.6\columnwidth]{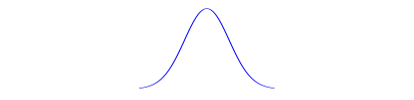}
  \includegraphics[width=0.6\columnwidth]{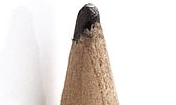}
  \caption{Perpendicular sections of a pencil line can be well approximated by a bell-like function (e.g. gaussian or arc of circumference). The trait intensity has a strong correlation with the pencil tip that traced it.}
  \label{fig:pencil}
\end{figure}

The most difficult of these tool traits to be robustly recognized is, by far, pencil. Unlike ink, pens and PC drawing, it presents a great ``hardness'' (color) variability.
Also, it is noisy and not constant along its perpendicular direction (bell shaped). Fig. \ref{fig:pencil} shows a simple demonstration of the reason of this. Moreover, artists may intentionally change the pressure while drawing to express artistic intentions.

\begin{figure}
  \centering
  \includegraphics[width=0.7\columnwidth]{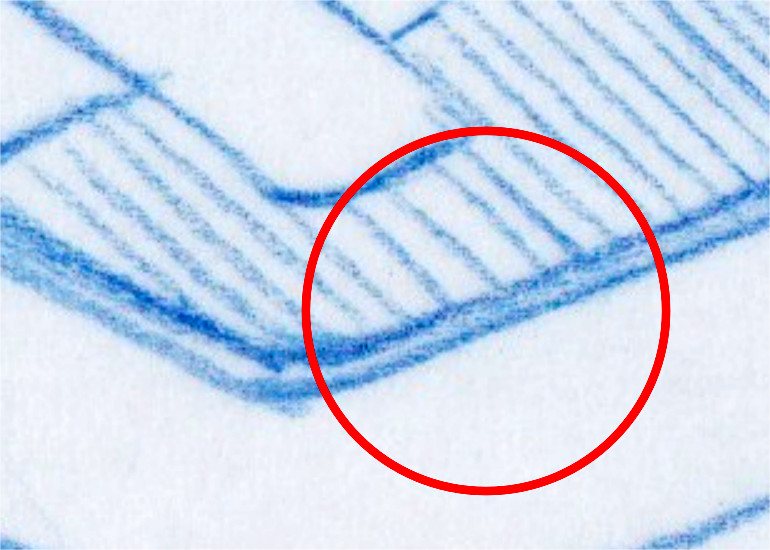}
  \caption{Detecting two parallel lines could be just a matter of stroke hardness and surrounding context.}
  \label{fig:hardness_circle}
\end{figure}

In addition, it is common for a ``wide'' line to be composed by multiple superimposed thinner traits. At the same time, parallel lines that should be kept separated may converge and almost touch in a given portion of the paint, having as the sole delimiter the brightness of the trait (Fig. \ref{fig:hardness_circle}).

With these premises, precise line extraction in this situation represents a great challenge. Our proposed approach is a custom line extraction mechanism that tries to be invariant to the majority of the foretold caveats, and aims to: be invariant to stroke hardness and stroke width; detect bell ``shaped'' lines, transforming them into classical ``uniform'' lines; merge multiple superimposed lines, while keeping parallel neighbor lines separated.

\subsubsection{Background about Cross Correlation}
The key idea behind the proposed algorithm is to exploit Pearson's Correlation Coefficient (PCC, hereinafter) and its properties to identify the parts of the image which resemble a ``line'', no matter the line width or strength. This section will briefly introduce the background about PCC.

The näive Cross Correlation is known for expressing the \emph{similarity} between two signals (or \emph{images} in the discrete 2D space), but it suffers of several problems, i.e. dependency on the sample average, the scale and the vector's sizes. In order to address all these limitations of Cross Correlation, Pearson's Correlation Coefficient (PCC) between two samples $a$ and $b$ can be used:
\begin{equation} \label{eq:pcc}
pcc(a,b) = \frac
			{cov(a, b)}
			{\sigma_a \sigma_b}
\end{equation}
where $cov(a, b)$ is the covariance between $a$ and $b$, and $\sigma_a$ and $\sigma_b$ are their standard deviations. From the definitions of covariance and standard deviation, eq. \ref{eq:pcc} can be re-written as follows:
\begin{equation} \label{eq:lin_inv}
    pcc(a,b) = pcc(m_0a + q_0, m_1b + q_1) \\
\end{equation}
${\forall q_{0,1} \land \forall m_{0,1}: m_0 m_1 > 0}$. Eq. \ref{eq:lin_inv} implies invariance to most affine transformations. Another strong point in favor of PCC is that its output value is of immediate interpretation. In fact, it holds that $ -1 \leq pcc(a, b) \leq 1$. $pcc \approx 1$ means that $a$ and $b$ are very correlated, whereas $pcc \approx 0$ means that they are not correlated at all.
On the other hand, $pcc \approx -1$ means that $a$ and $b$ are strongly inversely correlated (i.e., raising $a$ will decrease $b$ accordingly).

PCC has been used in the image processing literature and in some commercial machine vision applications, but mainly as an algorithm for object detection and tracking. Its robustness derives from the properties of illumination and reflectance, that apply to many real-case scenarios involving cameras. Since the main lighting contribution from objects is linear, $pcc$ will give very consistent results for varying light conditions, because of its affine transformations invariance (eq. \ref{eq:lin_inv}), showing independence from several real-world lighting issues. 

Stepping back to our application domain, at the best of our knowledge, this is the first paper proposing to use PCC for accurate line extraction from hand-drawn sketches. Indeed, PCC can grant us the robustness in detecting lines also under severe changes in the illumination conditions, for instance when images can potentially be taken from very diverse devices, such as a smartphone, a satellite, a scanner, an x-ray machine, etc.. Additionally, the ``source'' of the lines can be very diverse: from hand-drawn sketches, to fingerprints, to paintings, to corrupted textbook characters, etc.. In other words, the use of PCC makes our algorithm generalized and applicable to many different scenarios.

\subsubsection{Pearson's Correlation Coefficient applied to images}

In order to obtain the punctual PCC between an image $I$ and a (usually smaller) template $T$, for a given point $p = (x, y)$, the following equation can be used:
\begin{equation} 
\label{eq:punctual_pcc}
pcc(I, T, x, y) = \frac
			{\sum_{j, k}  \left(I_{xy}(j, k) - u_{I_{xy}}\right)\left(T\left(j, k\right) - u_T\right)}
			{\sqrt{\sum_{j, k} \left(I_{xy}(j, k) - u_{I_{xy}}\right)^2 \sum_{j, k} \left(T\left(j, k\right) - u_T\right)^2}}
\end{equation}
$\forall j \in [-T_w / 2; T_w / 2]$ and $\forall k \in [-T_h / 2; T_h / 2]$, and where $T_w$ and $T_h$ are the width and the height of the template $T$, respectively. $I_{xy}$ is a portion of the image $I$ with the same size of $T$ and centered around $p = (x, y)$. $u_{I_{xy}}$ and $u_T$ are the average values of $I_{xy}$ and $T$, respectively. $T(j, k)$ (and, therefore, $I_{xy}(j, k)$) is the pixel value of that image at the coordinates $j, k$ computed \emph{from the center} of that image.

It is possible to apply the punctual PCC from eq. \ref{eq:punctual_pcc} to all the pixels of the input image $I$ (except for border pixels). This process will produce a new image which represents how well each pixel of image $I$ resembles the template $T$. In the remainder of the paper, we will call it $PCC$. In Fig. \ref{fig:crossing_example} you see $PCC$s obtained with different templates. It is worth remembering that $PCC(x, y) \in [-1, 1], \forall 
x, y$. To perform just this computation, the input grayscale image has been inverted; in sketches usually lines are darker than white background, so inverting the colors gives us a more ``natural'' representation to be matched with a positive template/kernel.

\subsubsection{Template/Kernel for extracting lines}
\begin{figure}
\begin{center}
\subfloat[][]{\includegraphics[width=0.33\columnwidth]{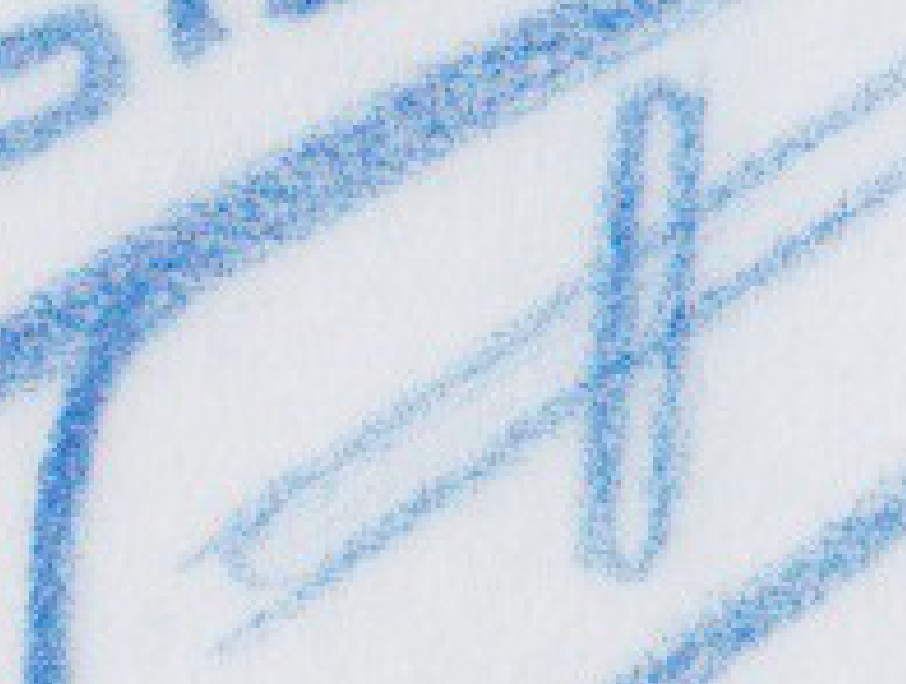}}
\subfloat[][]{\includegraphics[width=0.33\columnwidth]{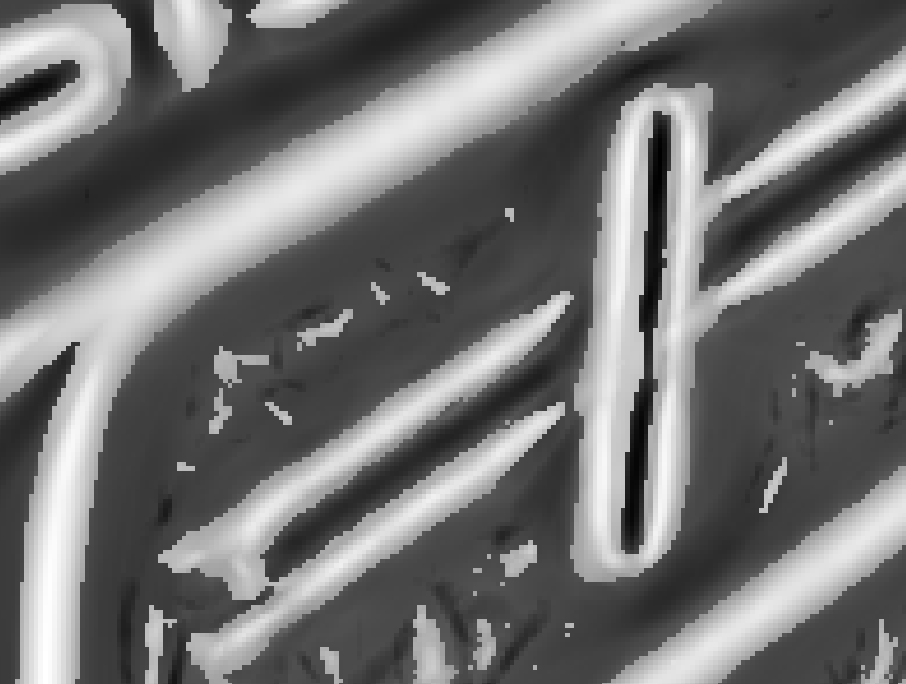}\label{fig:a2}}
\subfloat[][]{\includegraphics[width=0.33\columnwidth]{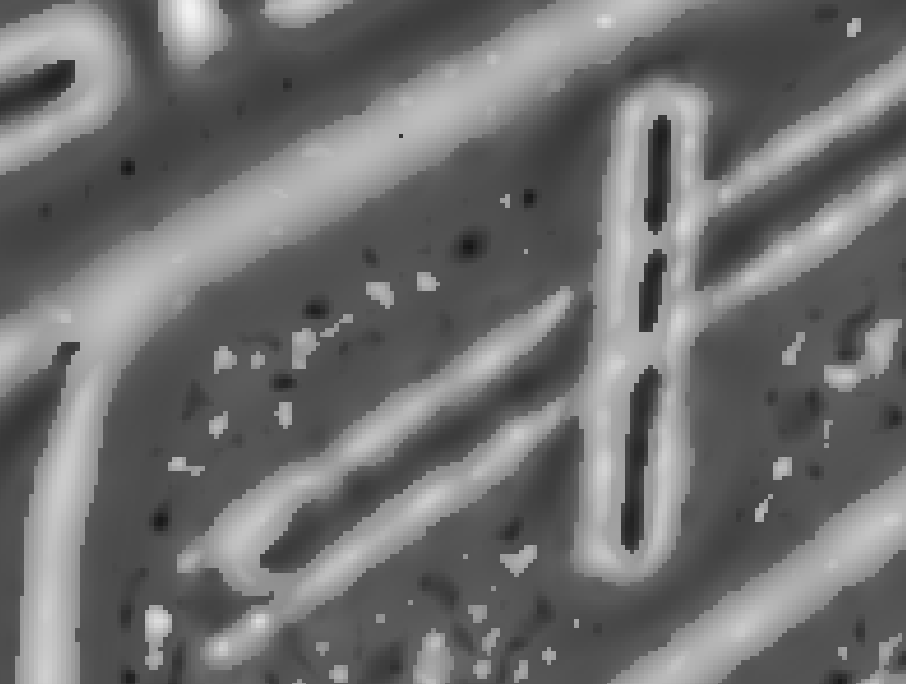}\label{fig:a3}}\\
\subfloat[][]{\includegraphics[width=0.33\columnwidth]{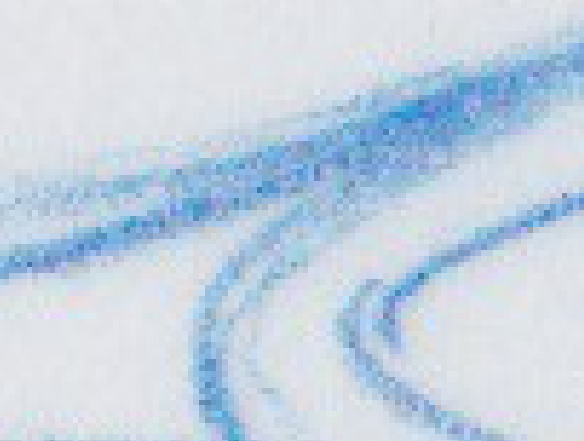}}
\subfloat[][]{\includegraphics[width=0.33\columnwidth]{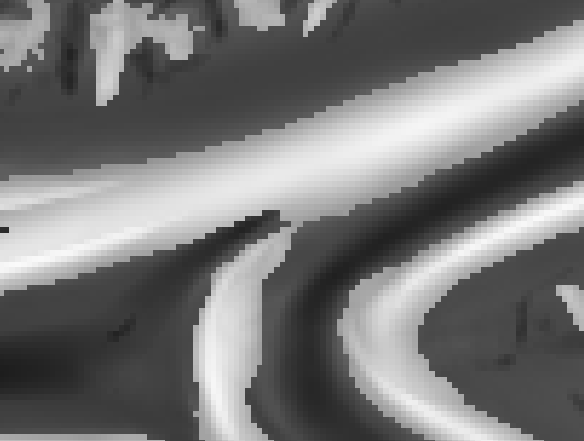}\label{fig:b2}}
\subfloat[][]{\includegraphics[width=0.33\columnwidth]{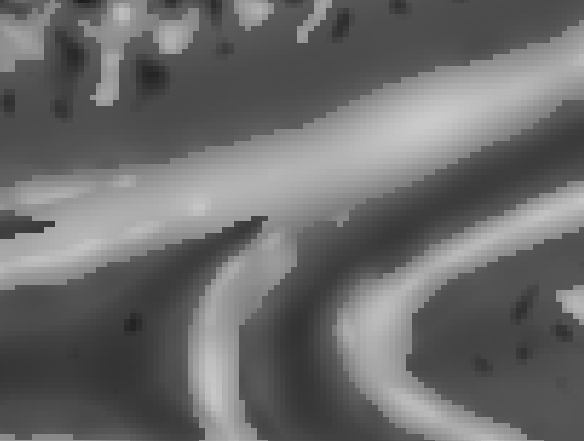}\label{fig:b3}}
\end{center}
\caption{Two examples of $PCC$ images obtained with different kernels. These pictures show that using a line-shaped kernel ($KLine$) can be detrimental for retrieval quality: \protect\subref{fig:a2}, \protect\subref{fig:b2}; crossing lines are truncated or detected as thinner than they should be. Using $KDot$ can alleviate the problem: \protect\subref{fig:a3}, \protect\subref{fig:b3}; this kernel detects more accurately ambiguous junctions.}
\label{fig:crossing_example}
\end{figure}

Our purpose is to extract lines from the input image. To achieve this, we apply $PCC$ with a suitable template, or \emph{kernel}. Intuitively, the best kernel to be used to find lines would be a sample approximating a ``generic'' line. A good generalization of a line might be a $1D$ Gaussian kernel replicated over the $y$ coordinate, i.e.:
$$KLine(x, y, \sigma) = gauss(x, \sigma)$$
This kernel achieves good detection results for simple lines, which are composed of clear (i.e., well separable from the background) and separated (from other lines) points. Unfortunately, this approach can give poor results in the case of multiple overlapping or perpendicularly-crossing lines. In particular, when lines are crossing, just the ``stronger'' would be detected around the intersection point. If both lines have about the same intensity, both lines would be detected, but with an incorrect width (extracted thinner than they should be).  An example is shown in the middle column of Fig. \ref{fig:crossing_example}. 

Considering these limitations, a full symmetric 2D Gaussian kernel might be more appropriate, also considering the additional benefit of being isotropic:
$$KDot(x, y, \sigma) = gauss(x, \sigma) \cdot gauss(y, \sigma)$$
This kernel has proven to solve the concerns raised with $KLine$, as shown in the rightmost column of Fig. \ref{fig:crossing_example}. In fact, this kernel resembles a dot, and considering a line as a continuous stroke of dots, it will approximate our problem just as well as the previous kernel. Moreover, it behaves better in line intersections, where intersecting lines become (locally) \emph{T-like} or \emph{plus-like} junctions, rather than simple straight lines. Unfortunately, this kernel will also be more sensitive to noise.

\subsubsection{Achieving size invariance}
One of the major objectives of this method is to detect lines without requiring finely-tuned parameters or custom ``image-dependent'' techniques. We also aim at detecting both small and large lines that might be mixed together, as it happens in many real drawings. In order to achieve invariance to variable line widths, we will be using kernels of different sizes.

We will generate $N$ Gaussian kernels, each with its $\sigma_i$. In order to find lines of width $w$ a sigma of $\sigma_i = w / 3$ would work, since a Gaussian kernel gives a contribution of about 84\% of samples at $3 \cdot \sigma$.

We follow an approach similar to the scale-space pyramid used in SIFT detector \cite{lowe1999object}. Given $w_{min}$ and $w_{max}$ as, respectively, the minimum and maximum line width to be detected, we can set $\sigma_0 = w_{min} / 3$ and ${\sigma_i = C\cdot\sigma_{i - 1} = C^i\cdot\sigma_0}$, $\forall i \in [1, N-1]$, where $N = log_C (w_{max} / w_{min})$, and $C$ is a constant factor or base (e.g., $C = 2$). Choosing a different base $C$ (smaller than 2) for the exponential and the logarithm will give a finer granularity.

The numerical formulation for the kernel will then be:
\begin{equation}
\footnotesize
KDot_i(x,y) = gauss(x - S_i / 2, \sigma_i) \cdot gauss(y - S_i / 2, \sigma_i)
\end{equation}
where $S_i$ is the kernel size and can be set as $Si = next\_odd(7\cdot\sigma_i)$, since the Gaussian can be well reconstructed in $7\cdot\sigma$ samples.

This generates a set of kernels that we will call $KDots$. We can compute the correlation image $PCC$ for each of these kernels, obtaining a set of images $PCCdots$, where ${PCCdots_i = pcc(Image, KDots_i)}$ with $pcc$ computed using eq. \ref{eq:punctual_pcc}.

\subsubsection{Merging results}
Once the set of images $PCCdots$ is obtained, we need to merge the results in a single image that can uniquely express the probability of line presence for a given pixel of the input image. This merging is obtained as follows:
\begin{equation} \label{eq:mpcc}
\footnotesize
    MPCC(x, y) =
    \begin{cases}
        maxPCC_{xy}, & \text{if } |maxPCC_{xy}| > |minPCC_{xy}| \\
        minPCC_{xy}, & \text{otherwise}
    \end{cases}
\end{equation}
\noindent where 
\begin{flushleft}
\begin{sloppypar}
${minPCC_{xy} = \min\limits_{\forall i \in [0, N-1]}PCCdots_i(x, y)}$,\\ ${maxPCC_{xy} = \max\limits_{\forall i \in [0, N-1]}PCCdots_i(x, y)}$.
\end{sloppypar}
\end{flushleft}

Given that $-1 \le pcc \le 1$ for each pixel, where $\approx 1 $ means strong correlation and $\approx -1 $ means strong inverse correlation, eq. \ref{eq:mpcc} tries to retain the most confident decision: ``it is definitely a line'' or ``it is definitely NOT a line''. 

By thresholding \emph{MPCC} of eq. \ref{eq:mpcc}, a binary image called $LinesRegion$ is obtained. The threshold has been set to 0.1 in all our experiments and resulted to be very stable in different scenarios. 

\subsubsection{Post-processing Filtering}
The binary image $LinesRegion$ will unfortunately still contain incorrect lines due to the random image noise. Some post-processing filtering techniques can be used, for instance, to remove too small connected components, or to delete those components for which the input image is too ``white'' (no strokes present, just background noise).

\begin{figure}
  \centering
  \includegraphics[width=1.0\columnwidth]{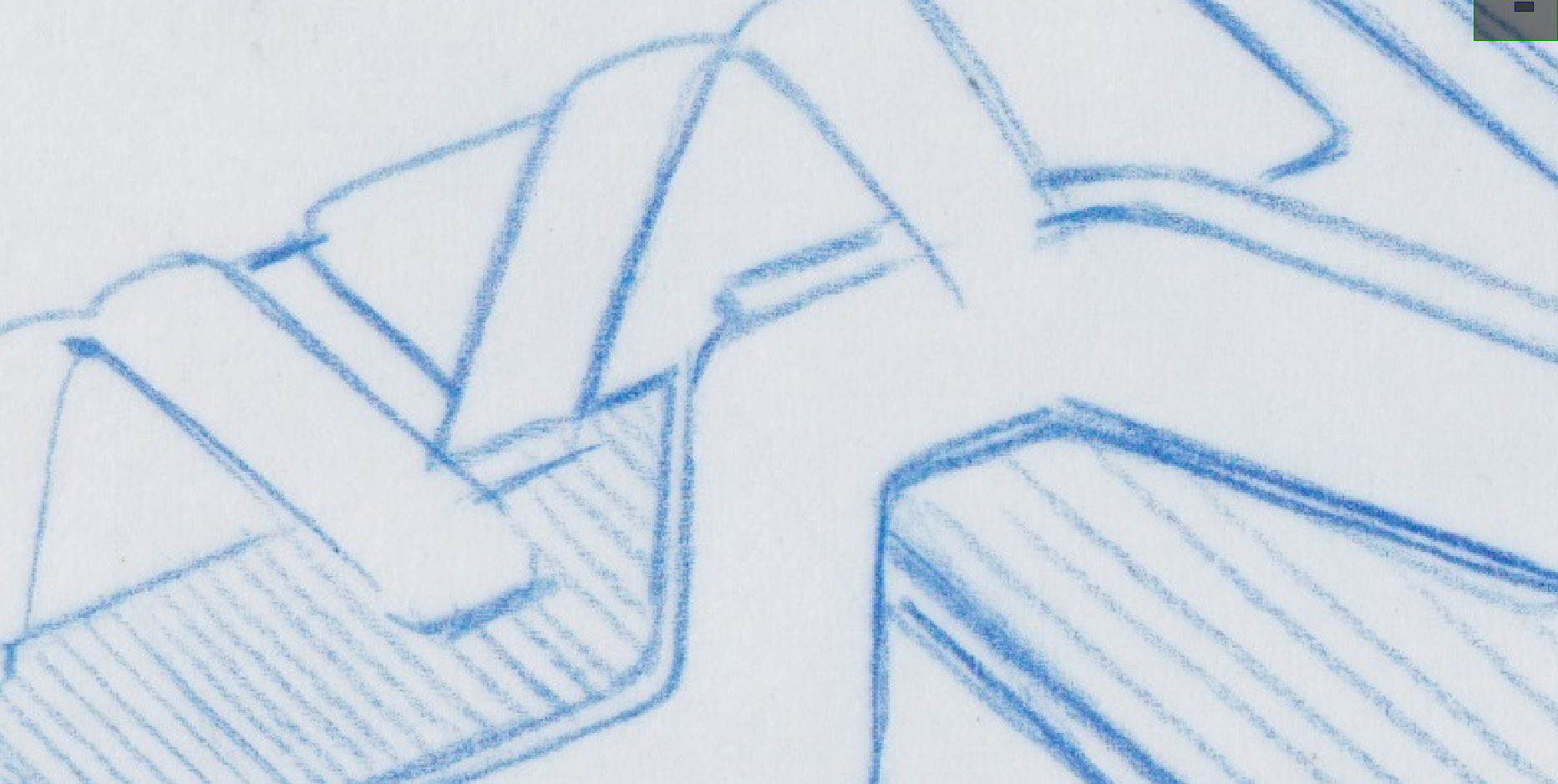}
  
  \smallskip
  
  \includegraphics[width=1.0\columnwidth]{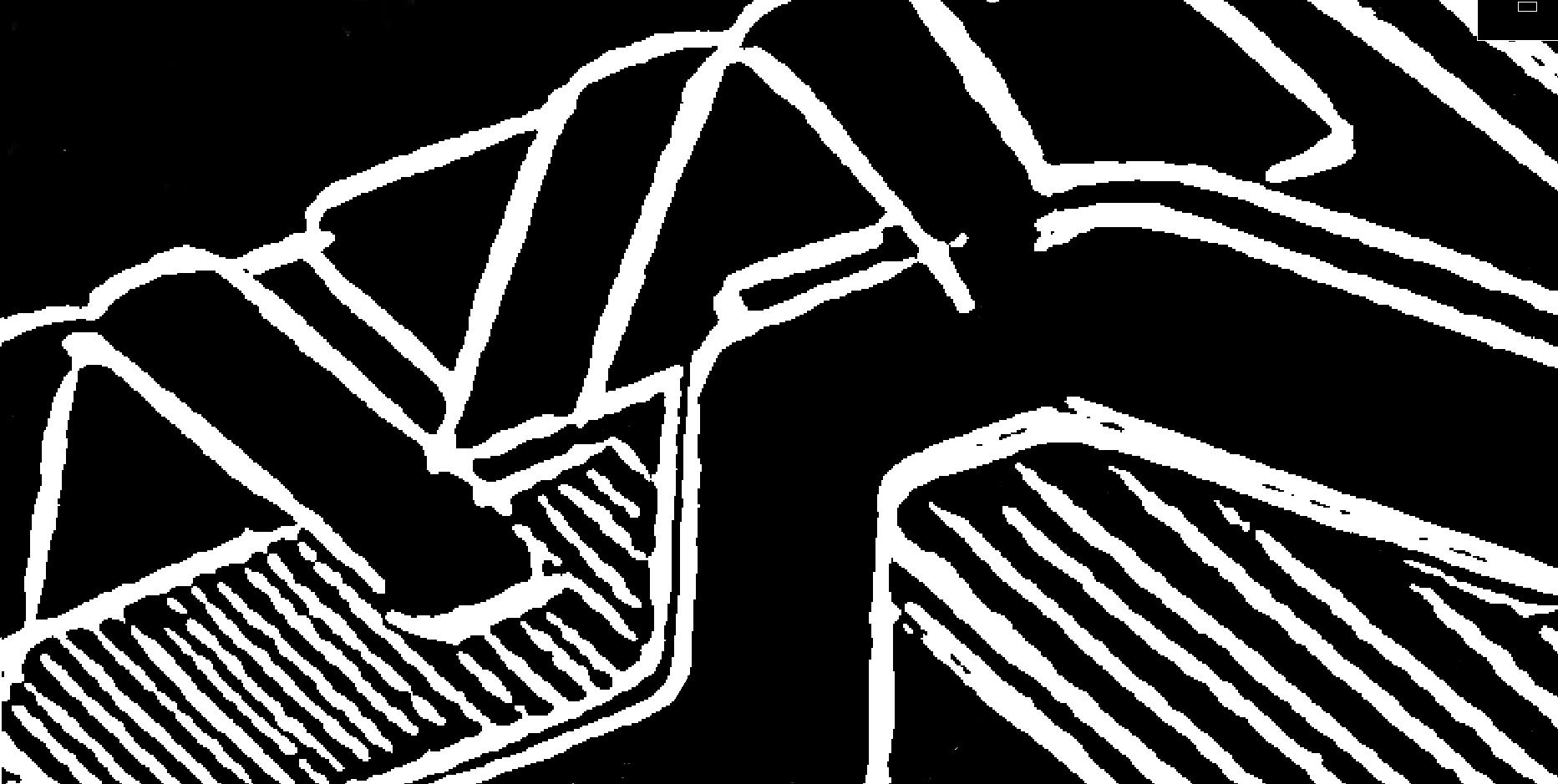}
  \caption{Part of a shoe sketch and its extracted $LinesRegion$ (after postprocessing).}
  \label{fig:linreg}
\end{figure}

For post-processing hand-drawn sketches, we first apply a high-pass filter to the original image, computing the median filter with window size $s > 2 \cdot w_{max}$ and subtracting the result from the original image value. Then, by using the well-known \cite{otsu1975threshold} method, the threshold that minimizes black-white intraclass variance can be estimated and then used to keep only the connected components for which the corresponding gray values are lower (darker stroke color) than this threshold. A typical output example can be seen in Fig. \ref{fig:linreg}.

\subsection{Thinning}
An extracted line shape is a ``clean'' binary image. After post-processing (holes filling, cleaning) it is quite polished. Still, each line has a varying, noisy width, and if we want to proceed towards vectorization we need a clean, compact representation. The natural choice for reducing line shapes to a compact form is to apply a thinning algorithm \cite{gonzalez2007image}.

Thinning variants are well described in the review \cite{KhalidSaeed2010}. In general terms, thinning algorithms can be classified in one-pass or multiple-passes approaches. The different approaches are mainly compared in terms of processing time, rarely evaluating the accuracy of the respective results. Since it is well-known, simple and extensively tested, we chose \cite{zhang1984fast}'s algorithm as baseline. However, any iterative, single-pixel erosion-based algorithm will work well for simple skeletonizations.

\begin{figure}
  \centering
  \includegraphics[width=0.48\columnwidth]{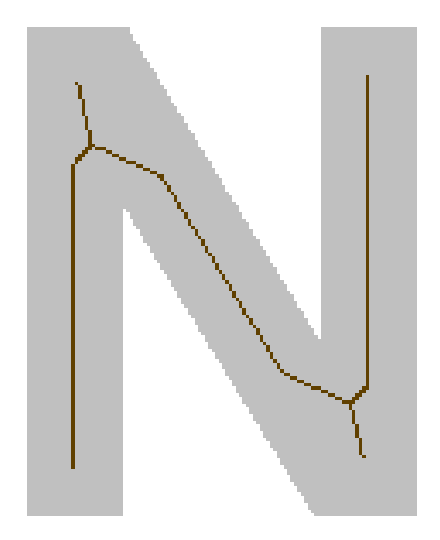}
  \includegraphics[width=0.48\columnwidth]{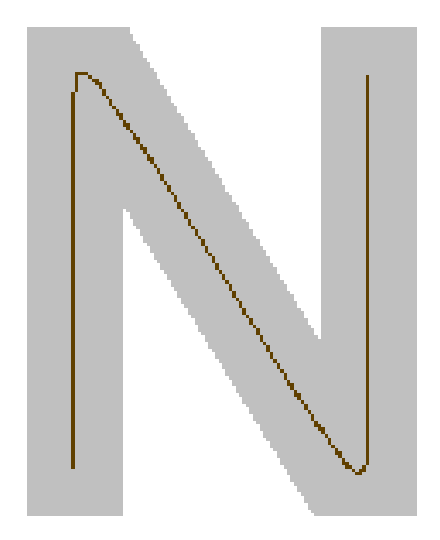}
  \caption{Example of the biasing effect while thinning a capital letter $N$ (on the left). On the right, the ideal representation of the shape to be obtained.}
  \label{fig:unbia}
\end{figure}

Unfortunately, Zhang and Suen's algorithm presents an unwanted effect known as ``skeletonization bias'' \cite{KhalidSaeed2010} (indeed, most of the iterative thinning algorithms produce biased skeletons). In particular, along steep angles the resulting skeleton may be wrongly shifted, as seen in Fig. \ref{fig:unbia}. The skeleton is usually underestimating the underlying curve structure, ``cutting'' curves too short. This is due to the simple local nature of most iterative, erosion-based algorithms. These algorithms usually work by eroding every contour pixel at each iteration, with the added constraint of preserving full skeleton connectivity (not breaking paths and connected components). They do that just looking at a local 8-neighborhood of pixels and applying masks. This works quite well in practice, and is well suited for our application, where the shapes to be thinned are lines (shapes already very similar to a typical thinning result). The unwanted bias effect arises when thinning is applied to strongly-concave angles. As described by the work of Chen \cite{chen1996use}, the bias effect appears when the shape to be thinned has a contour angle steeper (lower) than 90 degrees.

To eliminate this problem, we developed our custom unbiased thinning algorithm. The original proposal in \cite{chen1996use} is based, first, on the detection of the steep angles and, then, on the application of a ``custom'' local erosion specifically designed to eliminate the so-called ``hidden deletable pixels''. We propose a more rigorous method that generalizes better with larger shapes (where a 8-neighbors approach fails).

Our algorithm is based on this premise: a standard erosion thinning works equally in each direction, eroding one pixel from each contour at every iteration. However, if that speed of erosion (1 pixel per iteration) is used to erode regular portions of the shape, a faster speed should be applied at steep angle locations, if we want to maintain a well proportioned erosion for the whole object, therefore extracting a more correct representation of the shape.

\begin{figure}
  \centering
  \includegraphics[width=0.5\textwidth]{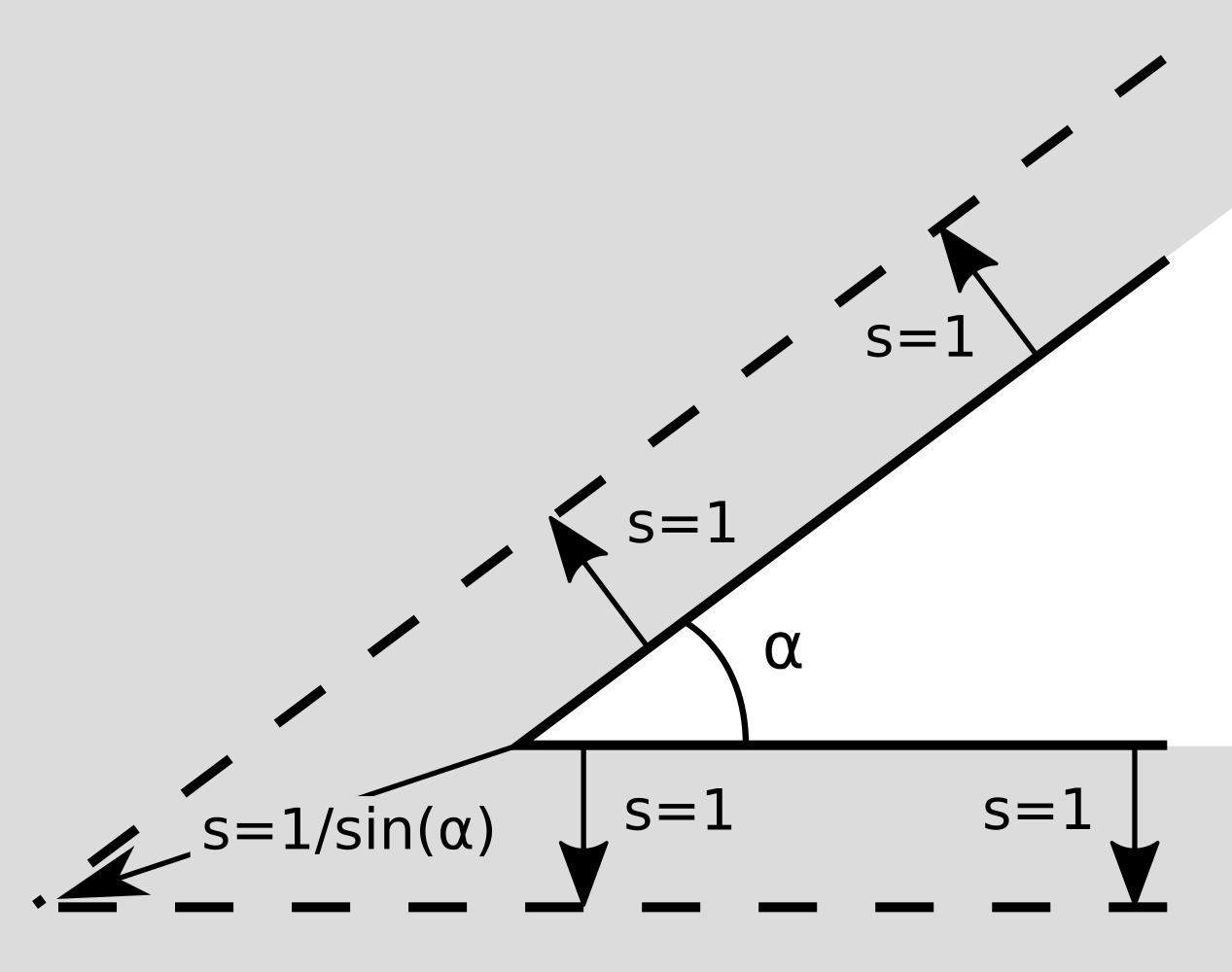}
  \caption{Applying an equal erosion to all the points of a concave shape implies eroding at a faster speed alongside steep angles. A speed of $s = 1/sin(\alpha)$ must be applied.
  }
  \label{fig:velocity}
\end{figure}

As shown in Fig. \ref{fig:velocity}, an erosion speed of:
$$ s = 1 / sin(\alpha)$$
should be applied at each angle point that needs to be eroded, where $\alpha$ is the angle size. Moreover, the erosion direction should be opposite to the angle bisector. In this way, even strongly concave shapes will be eroded uniformly over their whole contours.

\begin{figure}
\begin{center}
\begin{tikzpicture}
\matrix[square matrix]{
{1} & { } & { } \\
{ } & {1} & {0} \\
{ } & {0} & {1} \\
};
\end{tikzpicture}
\begin{tikzpicture}
\matrix[square matrix]{
{ } & {1} & { } \\
{ } & {1} & {0} \\
{ } & {0} & {1} \\
};
\end{tikzpicture}
\begin{tikzpicture}
\matrix[square matrix]{
{ } & {1} & { } \\
{0} & {1} & { } \\
{1} & {0} & { } \\
};
\end{tikzpicture}

\begin{tikzpicture}
\matrix[square matrix]{
{ } & {1} & { } \\
{0} & {1} & {0} \\
{ } & {1} & { } \\
};
\end{tikzpicture}
\begin{tikzpicture}
\matrix[square matrix]{
{ } & { } & { } \\
{0} & {1} & {0} \\
{1} & {0} & {1} \\
};
\end{tikzpicture}
\begin{tikzpicture}
\matrix[square matrix]{
{ } & {1} & { } \\
{1} & {1} & {1} \\
{ } & {1} & { } \\
};
\end{tikzpicture}
\end{center}
\caption{Four rotations of these masks of a hit-miss operator (1 = hit, 0 = miss, empty = ignored) are used to detect pixels necessary to preserve path connectivity.}
\label{fig:breaks}
\end{figure}
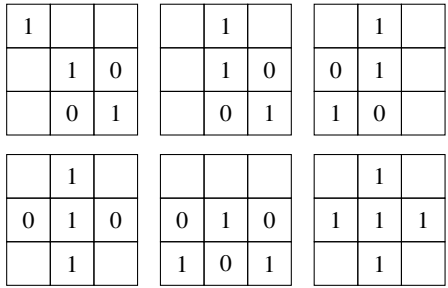

The steps of the algorithm are the following:
\begin{itemize}
\item First, we extract the contour of the shape to be thinned (by using the border-following algorithm described by \cite{suzuki1985topological}). This $contour$ is simply an array of 2d integer coordinates describing the shape outlines. Then, we estimate the curvature (angle) for each pixel in this contour. We implemented the technique proposed in \cite{han2001chord}, based on cord distance accumulation. Their method estimates the curvature for each pixel of a contour and grants good generalization capabilities. Knowing each contour pixel supposed curvature, only pixels whose angle is steeper than 90 degrees are considered. To find the approximate angle around a point of a contour the following formula is used:
$$\alpha \approx 6I_L / L^2$$
where $I_L$ is the distance accumulated over the contours while traveling along a cord of length $L$. Fig. \ref{fig:2} shows an example where concave angles are represented in green, convex in red, and straight contours in blue. Han \etal's method gives a rough estimate of angle intensity, but does not provide its direction. To retrieve it, we first detect the point of local maximum curvature, called $P_E$. Starting from it, the contours are navigated in the left direction, checking curvature at each pixel, until we reach the end of a straight portion of the shape (zero curvature - blue in Fig. \ref{fig:2}), which presumably concludes the angular structure (see Fig. \ref{fig:unbia_process}). This reached point is called $P_L$, the left limit of the angle. We do the same traveling right along the contour, reaching the point that we call $P_R$. These two points act as the angle surrounding limits.
\begin{figure*}
  \centering
  \subfloat[][]{\includegraphics[width=0.21\textwidth]{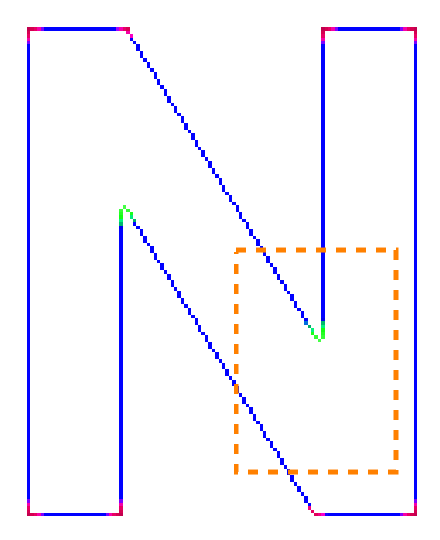}\label{fig:2}}\quad
  \subfloat[][]{\includegraphics[width=0.22\textwidth]{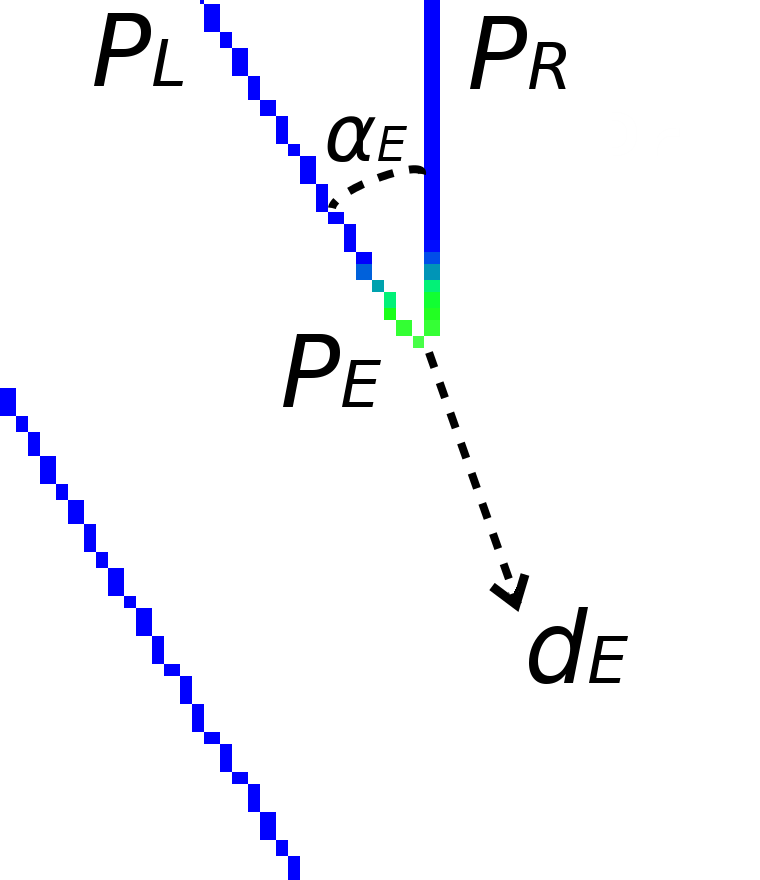}\label{fig:p1}}\\
  \subfloat[][]{\includegraphics[width=0.22\textwidth]{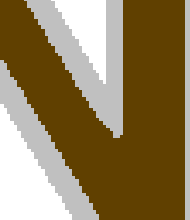}\label{fig:p2}}\quad
  \subfloat[][]{\includegraphics[width=0.22\textwidth]{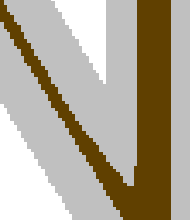}\label{fig:p3}}\quad
  \subfloat[][]{\includegraphics[width=0.22\textwidth]{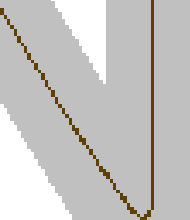}\label{fig:p4}}
  \caption{\protect\subref{fig:2} Curvature color map for the contours. Concave angles (navigating the contour clockwise) are presented in green, convex angles in red. Blue points are zero-curvature contours. \protect\subref{fig:p1} A portion of the contours curvature map, and a real example of the erosion-thinning process steps \protect\subref{fig:p2}, \protect\subref{fig:p3}, \protect\subref{fig:p4}. The green part of \protect\subref{fig:p1} is the neighborhood $N_{P_E}$ that will be eroded at the same time along $d_E$ direction.}
  \label{fig:unbia_process}
\end{figure*}
\item We then estimate the precise 2D direction of erosion and speed at which the angle point should be eroded. Both values can be computed by calculating the angle between segment $\overline{P_L P_E}$ and segment $\overline{P_E P_R}$, that we call $\alpha_E$. As already said, the direction of erosion $d_E$ is the opposite of $\alpha_E$ bisector, while the speed is $s_E = 1 / sin(\alpha_E)$.

\item After these initial computations, the actual thinning can start. Both the modified faster erosion of $P_E$ and the classical iterative thinning by \cite{zhang1984fast} are run in parallel. At every classical iteration of thinning (at speed $s=1$), the point $P_E$ is moved along its direction $d_E$ at speed $s_E$, eroding each pixel it encounters on the path. The fact that $P_E$ is moved at a higher speed compensates for the concaveness of the shape, therefore performing a better erosion of it. Figs. \ref{fig:p2}, \ref{fig:p3} and \ref{fig:p4} show successive steps of this erosion process.

\item Additional attention should be posed to not destroy the skeleton topology; as a consequence, the moving $P_E$ erodes the underlying pixel only if it does not break surrounding paths connectivity. Path connectivity is checked by applying four rotated masks of the hit-miss morphological operator, as shown in Fig. \ref{fig:breaks}. If the modified erosion encounters a pixel which is necessary to preserve path connectivity, the iterations for that particular $P_E$ stop for the remainder of the thinning.
\end{itemize}
To achieve better qualitative results, the faster erosion is performed not only on the single $P_E$ point, but also on some of its neighbor points (those who share similar curvature). We call this neighborhood set of points $N_{P_E}$ and are highlighted in green in Fig. \ref{fig:p1}. Each of these neighbor points $P_i$ should be moved at the same time with appropriate direction, determined by the angle $\alpha_i$ enclosed by segments $\overline{P_L P_i}$ and $\overline{P_i P_R}$. In this case, it is important to erode not only $P_i$, but also all the pixels that connect it (in straight line) with the next $P_{i+1}$ eroding point. This is particularly important because neighbor eroding pixels will be moving at different speeds and directions, and could diverge during time.

As usual for thinning algorithms, thinning is stopped when, after any iteration, the underlying skeleton has not changed (reached convergence).

\subsection{Creating and improving paths} \label{ssec:paths}
\subsubsection{Path creation}
The third major step towards vectorization is transforming the thinned image (a binary image representing very thin lines) in a more condensed and hierarchical data format. A good, simple representation of a thinned image is the set of paths contained in the image (also called ``contours''). A $path$ is defined as an array of consecutive 2D integer points representing a single ``line'' in the input thinned image:
$$ path = [(x_0, y_0), (x_1, y_1), ... (x_{n-1}, y_{n-1})] $$
In this paper, sub-pixel accuracy is not considered. A simple way to obtain sub-pixel accuracy could be zooming the input image before thinning and extracting real valued coordinates.

\begin{figure}
\begin{center}
\begin{tikzpicture}
\matrix[square matrix]{
{ } & {1} & { } \\
{ } & {1} & {1} \\
{0} & { } & { } \\
};
\end{tikzpicture}
\end{center}
\caption{Four rotations of this simple mask of \emph{in-place} hit-miss morph-operator (1 = hit, 0 = miss, empty = ignored) are used to transform a thinned image to a ``strictly 8-connected'' one.}
\label{fig:strictly}
\end{figure}
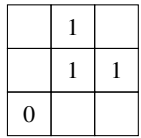

The successive steps of skeleton analysis will rely on 8-connectivity, so we need to convert the output of our thinning algorithm to that format. Our algorithm, being based on Zhang-Suen's thinning, produces 4-connected skeletons (with lines wide 1-2 pixels). A 4-connected skeleton can be transformed into a 8-connected one  by applying the four rotations of the mask shown in Fig. \ref{fig:strictly}. Pixels that match that mask must be deleted in the original skeleton (\emph{in-place}). Applying an \emph{in-place} hit-miss operator implies reading and writing consecutively on the same input skeleton.

The resulting thinned image is called a ``strictly 8-connected skeleton'': this definition does not imply that a pixel can not be connected with its 4-neighbors, but that each 4-neighbor pixel has been erased if not needed for topology preservation.

\subsubsection{Path classification}
\begin{figure}
\begin{center}
\begin{tikzpicture}
\matrix[square matrix]{
{ } & {1} & { } \\
{0} & {1} & {0} \\
{1} & {0} & {1} \\
};
\end{tikzpicture}
\begin{tikzpicture}
\matrix[square matrix]{
{1} & {0} & { } \\
{0} & {1} & {0} \\
{1} & {0} & {1} \\
};
\end{tikzpicture}
\begin{tikzpicture}
\matrix[square matrix]{
{ } & {0} & {1} \\
{1} & {1} & {0} \\
{ } & {1} & { } \\
};
\end{tikzpicture}
\end{center}
\caption{Four rotations of these masks of a hit-miss operator (1 = hit, 0 = miss, empty = ignored) are used to detect all the junctions in a ``strictly 8-connected skeleton''.}
\label{fig:junction}
\end{figure}
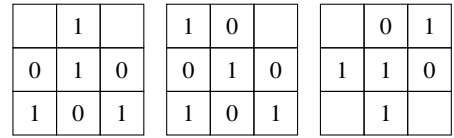

Once the ``strictly 8-connected skeleton'' is obtained, the path classification step is issued. In order to better understand how this step works, let us define some basic concepts. We first define a $junction$ between paths. A junction is a pixel that connects three or more paths. By deleting a junction pixel, paths that were connected are then split. We can detect all the junctions in the strictly 8-connected skeleton by applying the masks in Fig. \ref{fig:junction} (and their rotations) in a hit-miss operator. In this case, the algorithm must be performed with an input image (read only) and an output image (write only) and the order of execution does not matter (i.e., it can be executed in parallel).

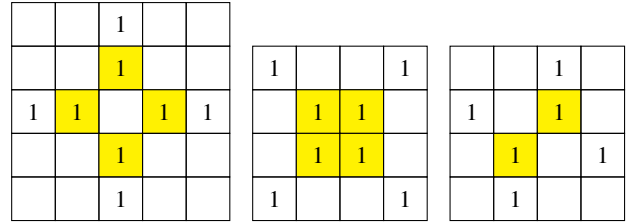
\begin{figure}
\newcommand{\jun}[0] {
  |[fill=yellow]| 1}
\begin{tikzpicture}
\matrix[square matrix]{
{ } & { } & {1} & { } & { } \\
{ } & { } & \jun & { } & { } \\
{1} & \jun & { } & \jun & {1} \\
{ } & { } & \jun & { } & { } \\
{ } & { } & {1} & { } & { } \\
};
\end{tikzpicture}
\begin{tikzpicture}
\matrix[square matrix]{
{1} & { } & { } & {1} \\
{ } & \jun & \jun & { } \\
{ } & \jun & \jun & { } \\
{1} & { } & { } & {1} \\
};
\end{tikzpicture}
\begin{tikzpicture}
\matrix[square matrix]{
{ } & { } & {1} & { } \\
{1} & { } & \jun & { } \\
{ } & \jun & { } & {1} \\
{ } & {1 } & { } & { } \\
};
\end{tikzpicture}
\caption{Examples of adjacent junctions (highlighted). Each of these junctions can not be deleted without changing the underlying topology, but can be treated as one.}
\label{fig:multijunctions}
\end{figure}

Sometimes, for peculiar input conformations, a junction pixel could be neighbor of another junction (see Fig. \ref{fig:multijunctions}). These specific junctions are merged together for successive analysis. Each path connecting to one of these junctions is treated as connecting to all of them. For simplicity, we choose one of them as representative.

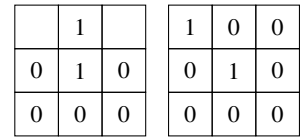
\begin{figure}
\begin{center}
\begin{tikzpicture}
\matrix[square matrix]{
{ } & {1} & { } \\
{0} & {1} & {0} \\
{0} & {0} & {0} \\
};
\end{tikzpicture}
\begin{tikzpicture}
\matrix[square matrix]{
{1} & {0} & {0} \\
{0} & {1} & {0} \\
{0} & {0} & {0} \\
};
\end{tikzpicture}
\end{center}
\caption{Four rotations of these masks of a hit-miss operator (1 = hit, 0 = miss, empty = ignored) are used to detect all the endpoints in a ``strictly 8-connected skeleton''.}
\label{fig:endpoint}
\end{figure}

Similarly, an $endpoint$ of a path can be defined as a pixel connected with one and only one other pixel, which needs to be part of the same path. Basically, an endpoint corresponds to either the starting or the ending point of a path. Endpoints can be straightforwardly found by applying the hit-miss morphological operator with the rotated versions of the masks reported in Fig. \ref{fig:endpoint}.

Given these definitions and masks, we can detect all the endpoints and the junctions of the strictly 8-connected skeleton. Then, starting from each of these points and navigating the skeletons, we will at the end reach other endpoints or junctions. By keeping track of the traversed pixels we can compose paths, grouped in one of these categories:
\begin{itemize}
\item paths from an endpoint to another endpoint ($e \leftrightarrow e$);
\item paths from a junction to endpoint ($j \leftrightarrow e$);
\item paths from a junction to a junction ($j \leftrightarrow j$).
\end{itemize}
There exists one last type of path that can be found, i.e. the closed path. Obviously, a closed path has no junctions or endpoints. In order to detect closed paths, all the paths that cannot be assigned to one of the three above groups are assigned to this category.

\begin{figure}
  \centering
  \subfloat[][]{\includegraphics[width=0.24\textwidth]{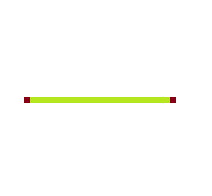}\label{fig:ee}}
  \subfloat[][]{\includegraphics[width=0.24\textwidth]{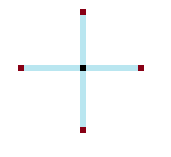}\label{fig:ej}}

  \subfloat[][]{\includegraphics[width=0.24\textwidth]{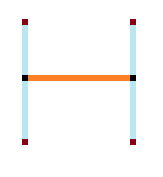}\label{fig:jj}}
  \subfloat[][]{\includegraphics[width=0.24\textwidth]{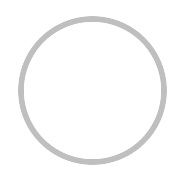}\label{fig:clo}}
  \caption{Examples of the four types of paths: $e \leftrightarrow e$ \protect\subref{fig:ee}, $e \leftrightarrow j$ \protect\subref{fig:ej}, $j \leftrightarrow j$ \protect\subref{fig:jj} and closed \protect\subref{fig:clo}. Junctions and endpoints are highlighted.}
  \label{fig:ptypes}
\end{figure}

\begin{figure}
  \centering
  \subfloat[][]{\includegraphics[width=0.24\textwidth]{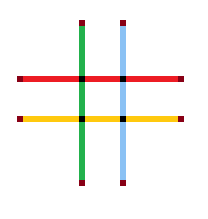}\label{fig:sa}}
  \subfloat[][]{\includegraphics[width=0.24\textwidth]{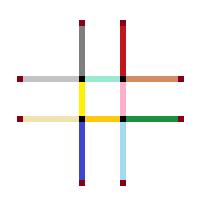}\label{fig:ttt}}
  \caption{Examples of semantic-aware connectivity \protect\subref{fig:sa}, where only four paths are detected, opposed to ``tic-tac-toe'' connectivity \protect\subref{fig:ttt}, where twelve small paths are detected. In this second case, overlapping paths are treated separately as different paths and no semantic knowledge of the context is retained.}
  \label{fig:conn}
\end{figure}

Examples of all types of paths are reported in Fig. \ref{fig:ptypes}. Fig. \ref{fig:conn} also shows a notable configuration, called ``tic-tac-toe'' connectivity. In this case, an assumpion is made, i.e. paths cannot overlap or cross each other. If this happens, paths are treated as distinct paths. In order to recognize that two crossing paths actually belong to the same ``conceptual'' path, a strong semantic knowledge of the underlying structures is necessary, but this is beyond the scope of this paper. A treatment of the problem can be found in \cite{Bo201614}, where they correctly group consecutive intersecting line segments.

\subsubsection{Path post-processing}
The resulting paths are easy to handle as they contain condensed information about the underlying shapes. However, they can often be further improved using some prior knowledge and/or visual clues from the original sketch image. Examples of these improvements are \emph{pruning}, \emph{merging} and \emph{linking}.

\begin{figure}
  \centering
  \subfloat[][]{\includegraphics[width=0.23\textwidth]{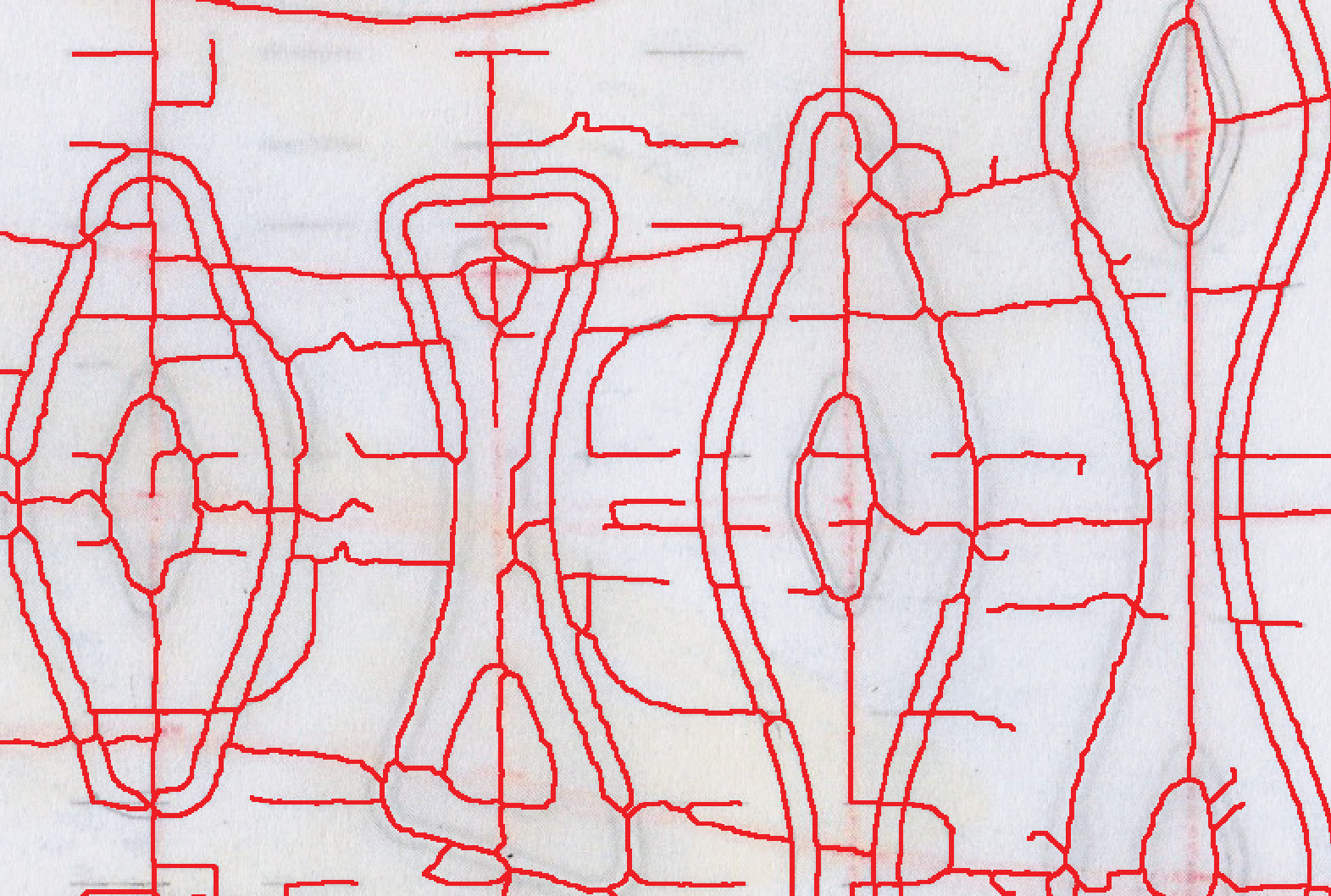}\label{fig:bp}}\enskip
  \subfloat[][]{\includegraphics[width=0.23\textwidth]{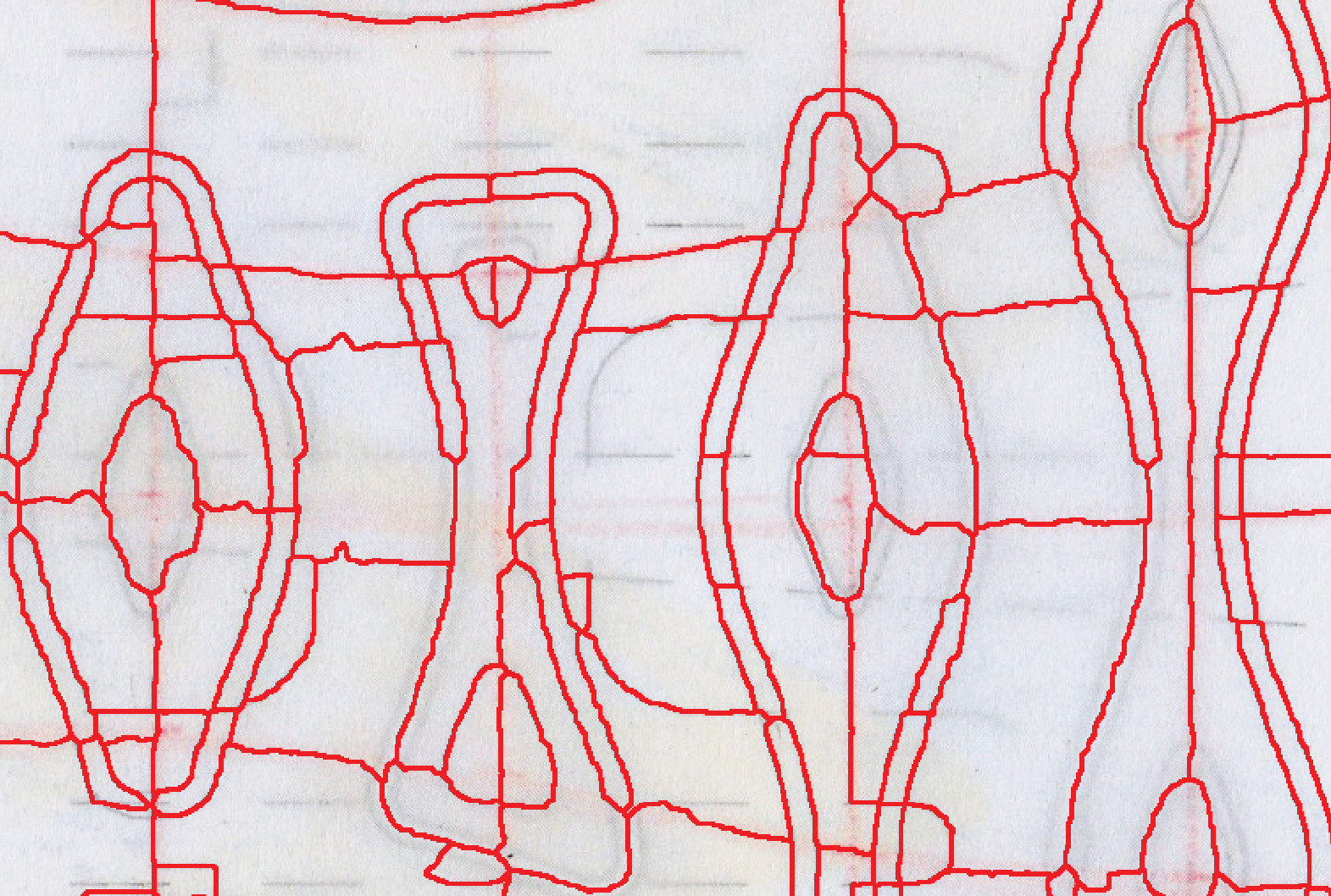}\label{fig:ap}}
  \caption{An example of $pruning$ applied to an input image \protect\subref{fig:bp}. Small branches are deleted from the resulting image \protect\subref{fig:ap}.}
  \label{fig:pru}
\end{figure}

\emph{Pruning} consists in deleting small, unwanted ``branches'' from the thinning results. The skeleton is usually composed by many long paths that contain most of the information needed. However, also smaller, unwanted ``branches'' are often present. They may be artifacts of thinning, or resulting from noise in the original image. By applying the pruning, branches shorter than a given length $l$ can be deleted. Branches are paths of the first or second type: $e \leftrightarrow e$ or $j \leftrightarrow e$. They can not belong to the third type, $j \leftrightarrow j$, because we do not want to alter skeleton connectivity by deleting them. An example of pruning is reported in Fig. \ref{fig:pru}. Pruning can be performed with different strategies, by deciding to keep more or less details. One simple idea is to iteratively prune all the image with increasing branch length threshold $l$.

\begin{figure}
  \centering
  \subfloat[][]{\includegraphics[width=0.23\textwidth]{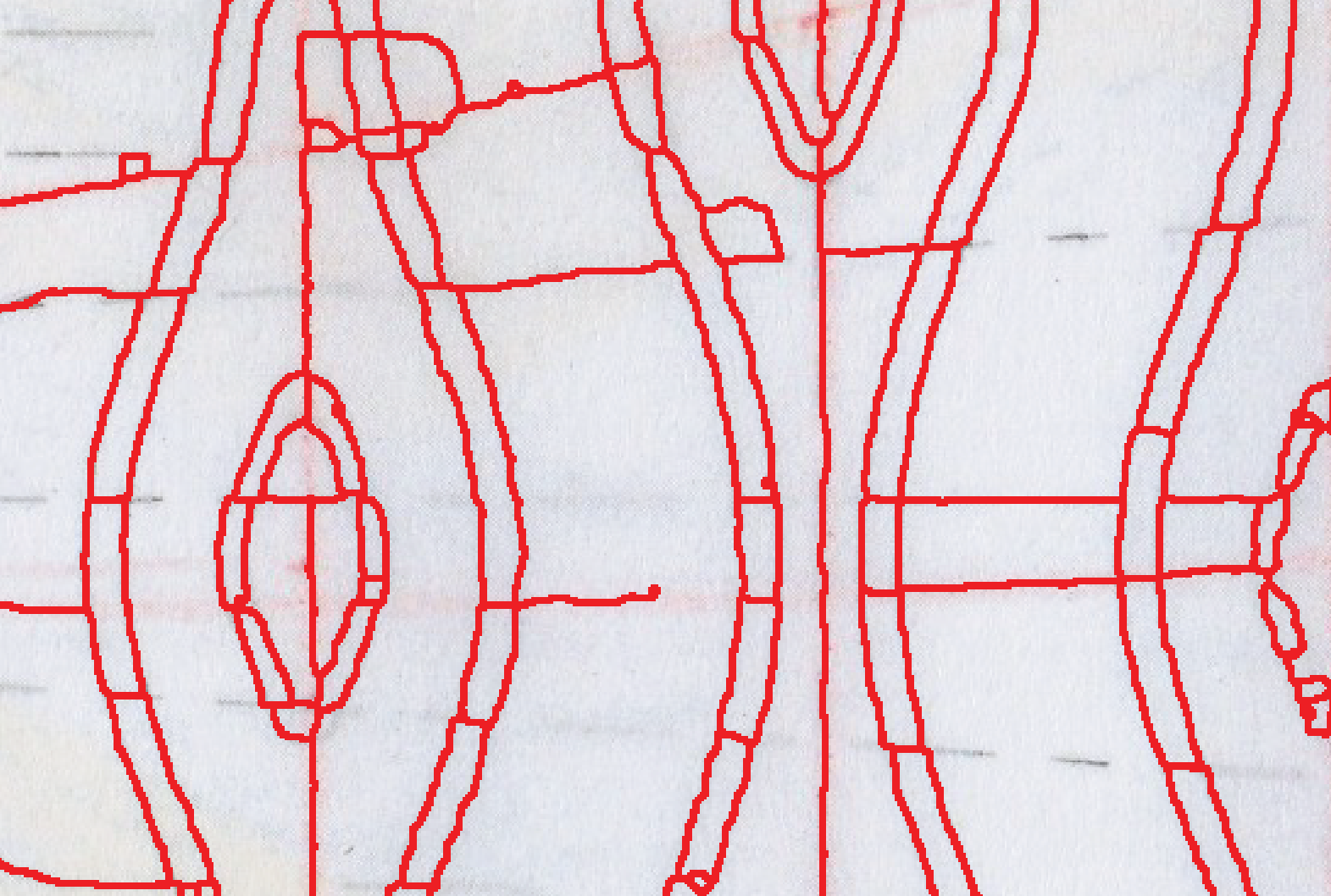}\label{fig:bm}}\enskip
  \subfloat[][]{\includegraphics[width=0.23\textwidth]{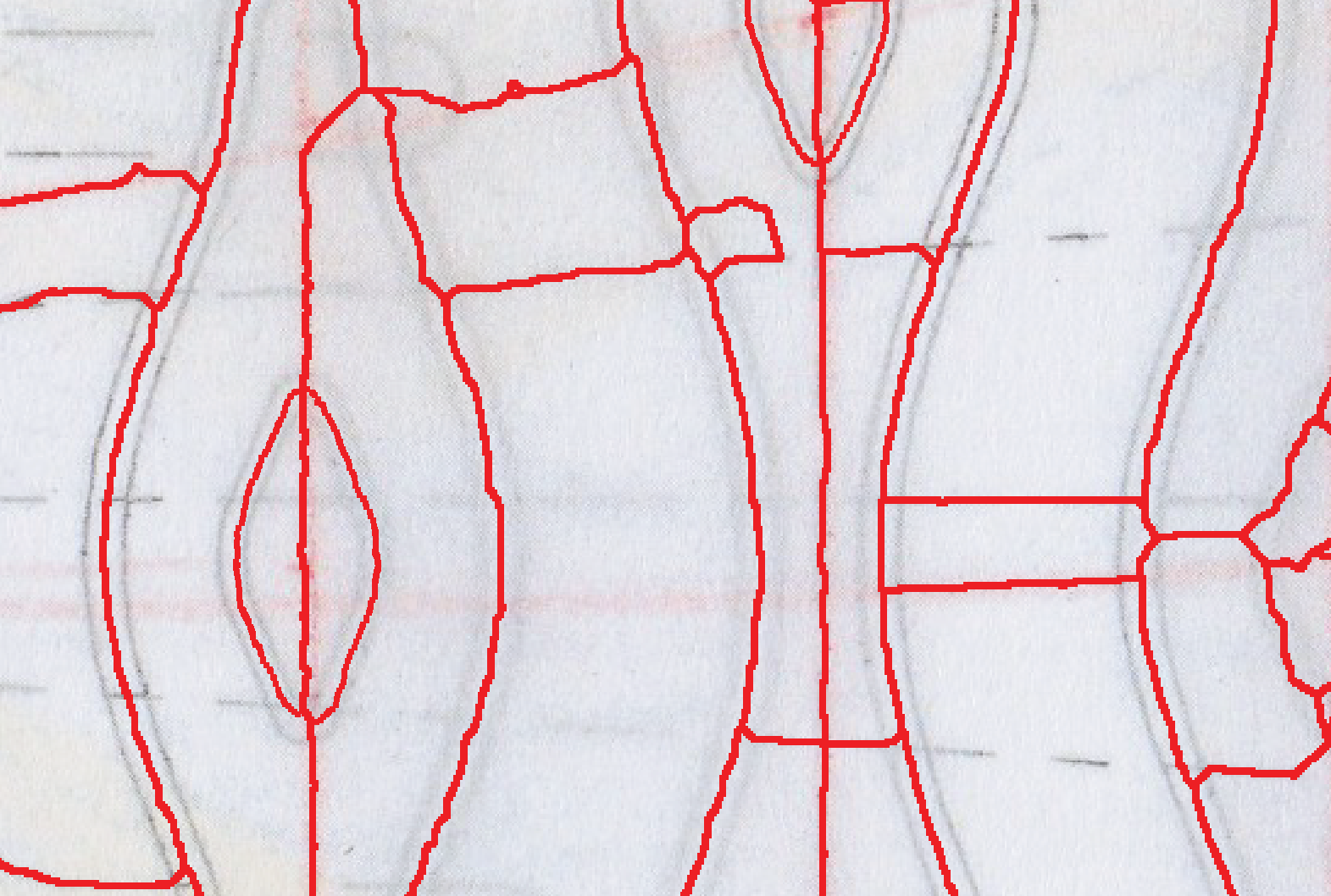}\label{fig:am}}
  \caption{An example of $merging$ applied to an input image \protect\subref{fig:bm}. Parallel paths are combined together in the resulting image \protect\subref{fig:am}.}
  \label{fig:mer}
\end{figure}

\emph{Merging} is the process of grouping together junctions or paths.  Junctions that are close each other could be grouped together in a single junction, simplifying the overall topology. After doing that, the same can be done for paths. Parallel and near paths that start and end in the same junctions are good candidates for merging (see an example in Fig. \ref{fig:mer}).

\begin{figure}
  \centering
  \subfloat[][]{\includegraphics[width=0.23\textwidth]{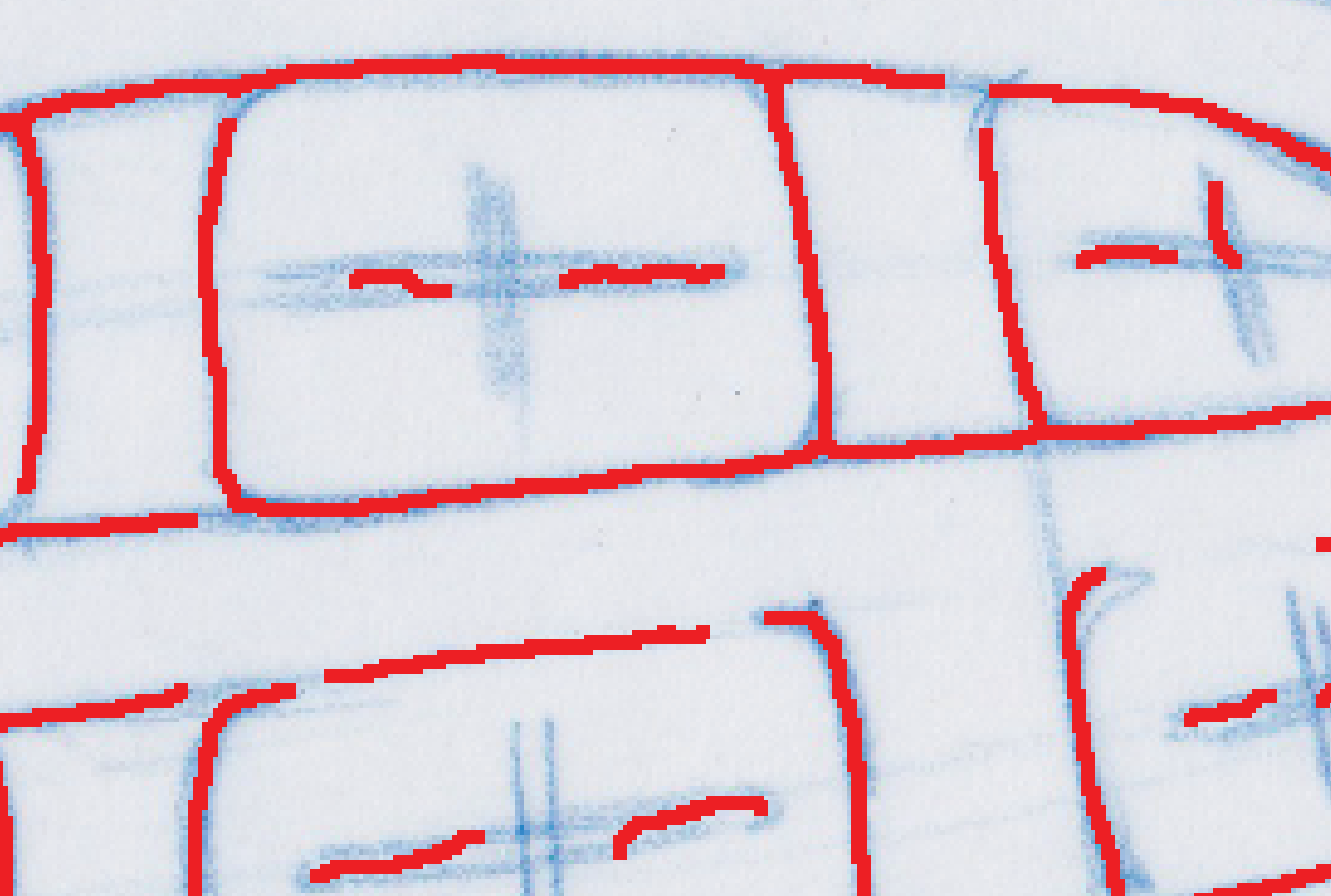}\label{fig:bl}}\enskip
  \subfloat[][]{\includegraphics[width=0.23\textwidth]{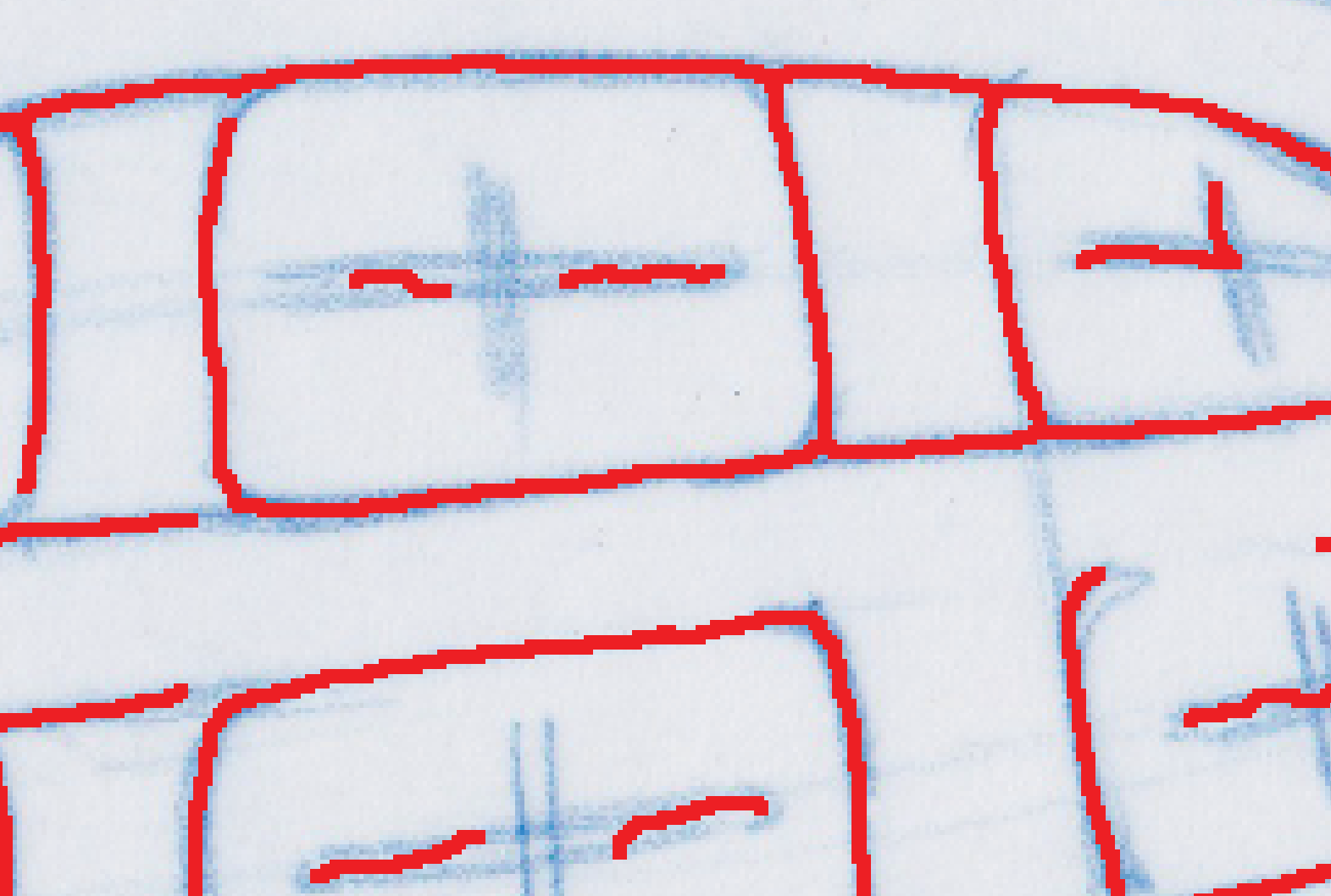}\label{fig:al}}
  \caption{An example of endpoints $linking$ applied to an input image \protect\subref{fig:bl}. Paths with adjacent endpoints have been connected in the output \protect\subref{fig:al}.}
  \label{fig:lin}
\end{figure}

\emph{Linking} (or endpoint linking, or edge linking) is the technique of connecting two paths whose endpoints are close in order to create a single path. Besides the endpoints distance, a good criteria for linking could be the path directions at their endpoints. Incident and near paths are suitable to be linked into a single, more representative, path (an example is reported in Fig. \ref{fig:lin}).

In order to improve their accuracy, all these post-processing techniques might benefit from the data (pixel values) from the original color image.

\subsection{Vectorization process}
Once cleaned paths are obtained, they need to be converted in a vectorized version with the minimum number of points. The basic vectorization algorithm used was firstly introduced in \cite{Schneider:1990:AAF:90767.90941}. Schneider's algorithm tries to solve a linear system (with least squares method), fitting a Bezier curve for each of the obtained paths. In detail, it tries iteratively to find the best set of cubic Bezier parameters that minimize the error, defined as the maximum distance of each path point from the Bezier curve. At each iteration, it performs a Newton-Raphson re-parametrization, to adapt the underlying path representation to the Bezier curve, in order to create a more suitable linear system to be optimized.

The algorithm is parametrized with a desired error to be reached, and a maximum number of allowed iterations. Whenever it converges, the algorithm returns the four points representing the best-fitting Bezier curve. If the convergence is not reached within the maximum number of iterations, the algorithm splits the path in two and recursively looks for the best fitting for each of them separately. The path is split around the point which resulted to have the maximum error w.r.t. the previous fitting curve. An additional constraint is related to C1 continuity of the two resulting curves on the splitting point, in order to be able to connect them smoothly.

In order to be faster, the original algorithm skips automatically all the curves which do not correspond to a minimum error (called ``iteration-error'' $\Psi$), and proceeds to the splitting phase without trying the Newton-Raphson re-parametrization. However, this simplification also affects accuracy of the vectorization, by generating a more complex representation due to the many split points created. Therefore, since the computational complexity is not prohibitive (worst case is $O(n\cdot \log n)$, with $n$ being the path length), we modified the original algorithm by removing this simplification.

Conversely, another early-stop condition has been introduced in our variant. Whenever the optimization reaches an estimation error lower than a threshold after having run at least a certain number of iterations, the algorithm stops and the estimated Bezier curve is returned. This can be summarized by the following condition:
\begin{lstlisting}[language=Python, breaklines=true, basicstyle=\small]
if current_err < desired_err * f and current_iter > tot_iter * f:
    return current_bezier
\end{lstlisting}
where ``f'' is an arbitrary fraction (set to $0.1$ in our experiments). This condition speeds up the algorithm if the paths are easily simplified (which is common in our case), while the full optimization process is run for ``hard'' portions that need more time to get good representations.

We also extended Schneider's algorithm to work for closed paths. First, C1 continuity is imposed for an arbitrary point of the path, selected as the first (as well as the last) point of the closed path. A first fit is done using the arbitrary point. Then, the resulting point with the maximum error w.r.t. the fitted curve is selected as a new first/last point of the path and the fitting algorithm is run a second time. In this way, if the closed path has to be fitted to two or more Bezier curves, the split point will be the one of highest discontinuity, not a randomly chosen one.

\section{Experiments}\label{sec:experiments}
\subsection{Line extraction}\label{line_ex_results}
In order to assess the accuracy of the proposed line extraction method, we performed an extensive evaluation on different types of images. First of all, we have used a large dataset of hand-drawn shoe sketches (courtesy of \financer). These sketches have been drawn from expert designers using different pens/pencils/tools, different styles and different backgrounds (thin paper, rough paper, poster board, etc.). Each image has its peculiar size and resolution, and has been taken from scanners or phone cameras.

\begin{figure}[bth]
\begin{center}
\subfloat[][]{\includegraphics[width=0.48\linewidth]{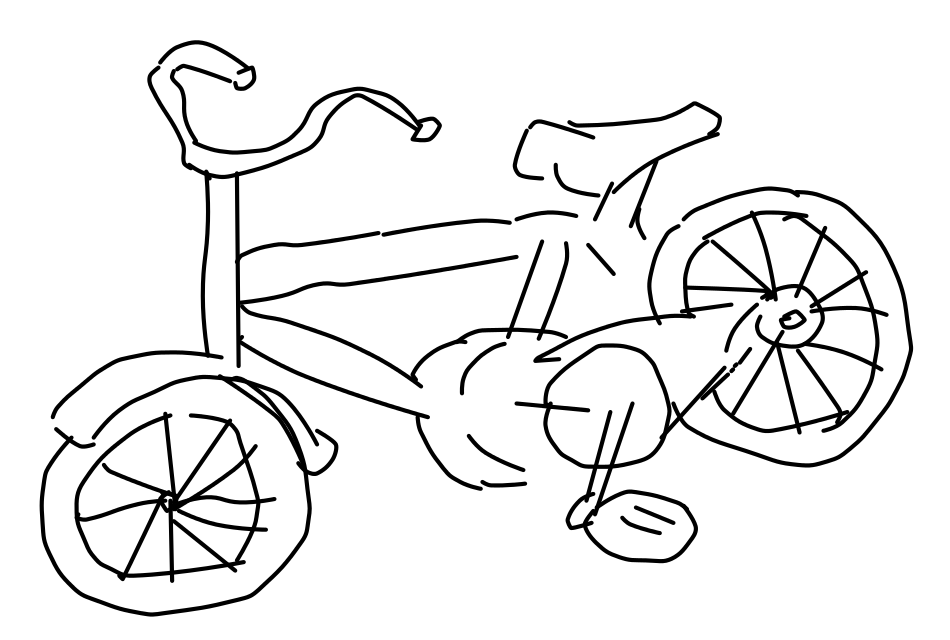}\label{fig:bike}}
\subfloat[][]{\includegraphics[width=0.48\linewidth]{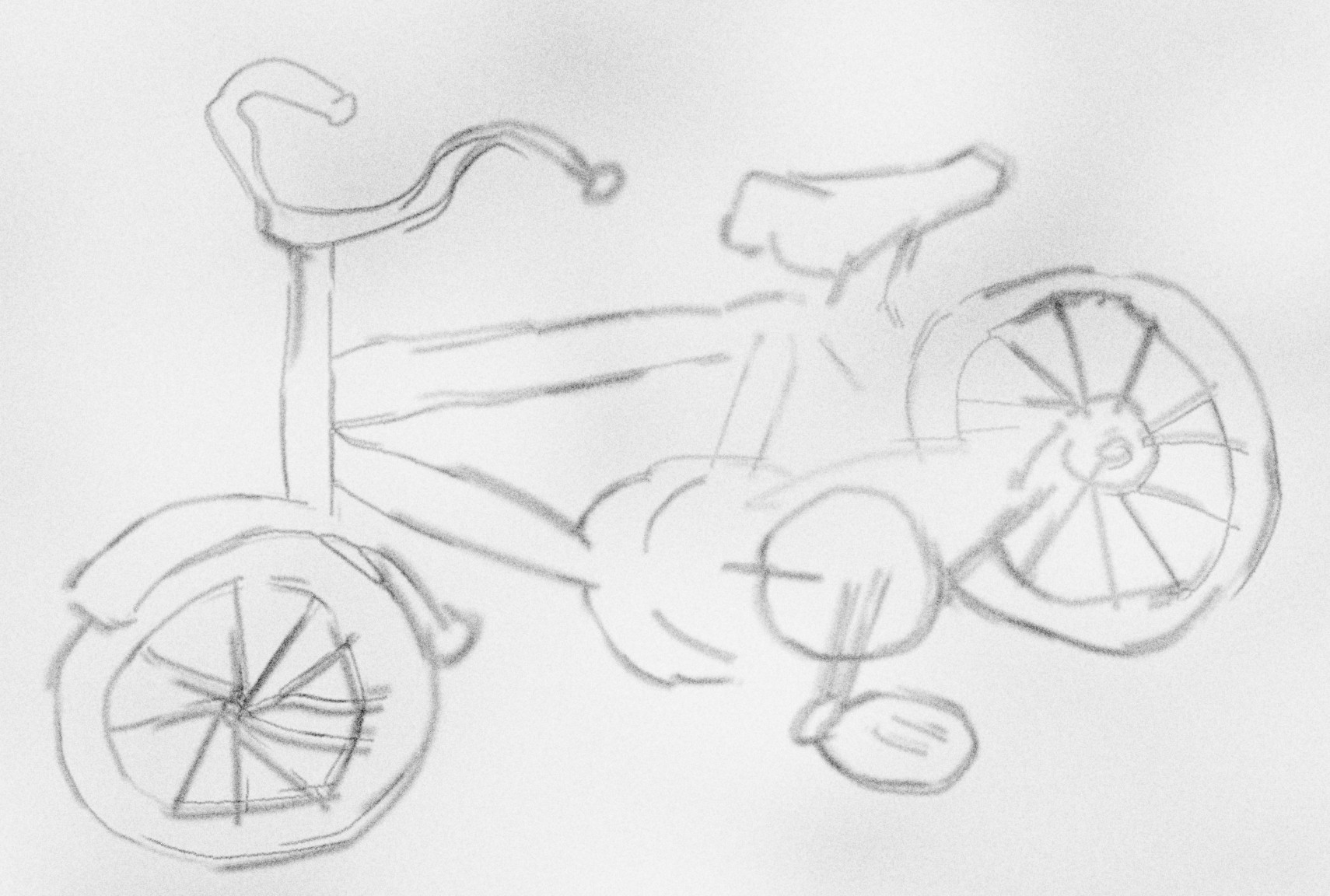}\label{fig:bike-after}}

\subfloat[][]{\includegraphics[width=0.48\linewidth]{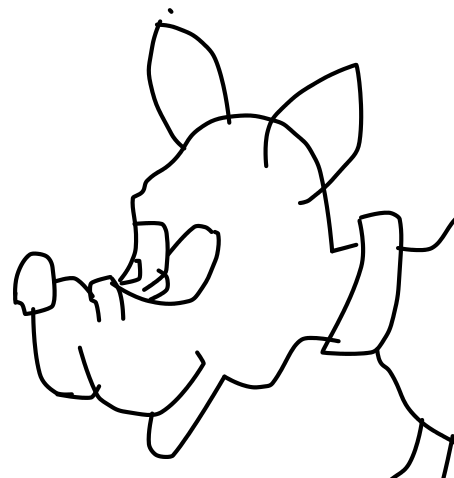}\label{fig:dogface}}
\subfloat[][]{\includegraphics[width=0.48\linewidth]{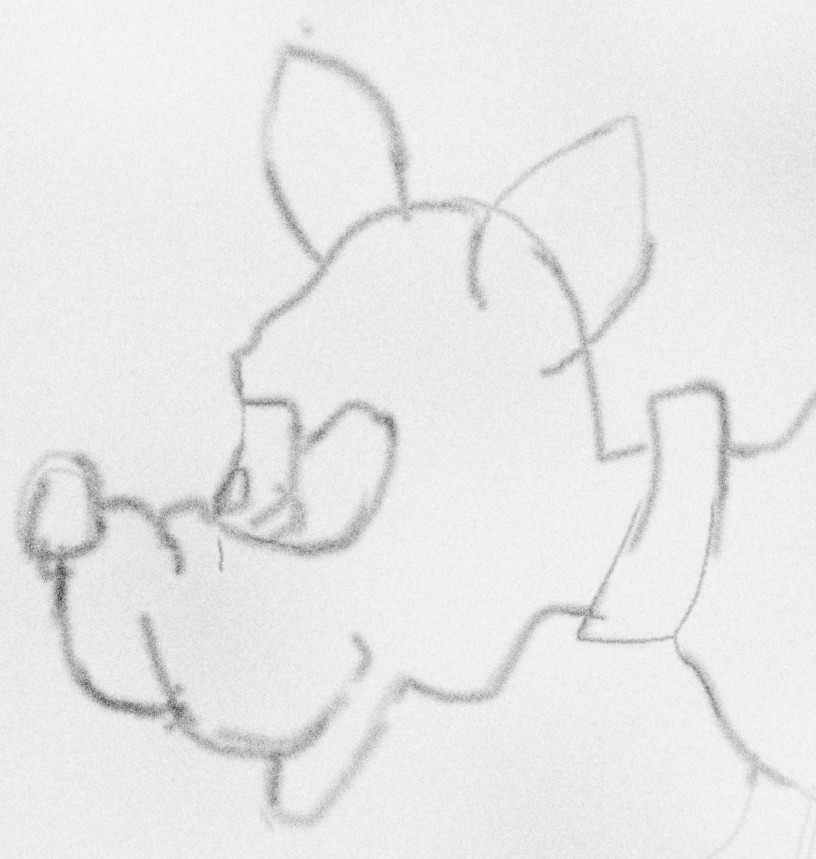}\label{fig:dogface-after}}
\end{center}
\caption{Examples of the ``inverse dataset'' sketches created from SHREC13 \cite{li2013shrec}.}
\label{fig:inverse}
\end{figure}

\begin{figure*}[bth]
\centering
\subfloat[][Input sketch]{\includegraphics[width=0.248\linewidth]{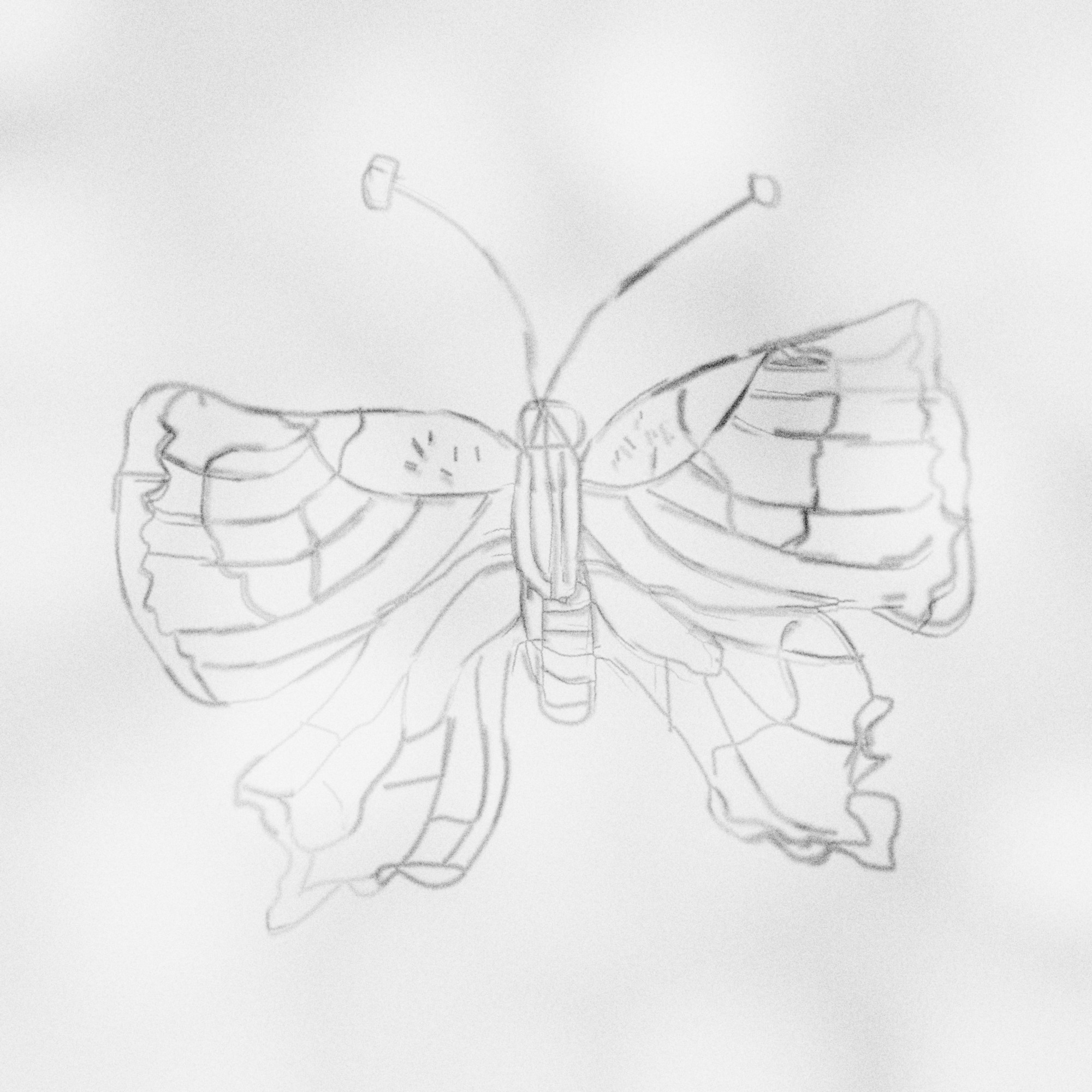} \label{fig:butterfly}}
\subfloat[][AI\textsuperscript{TM} Live Trace]{\includegraphics[width=0.248\linewidth]{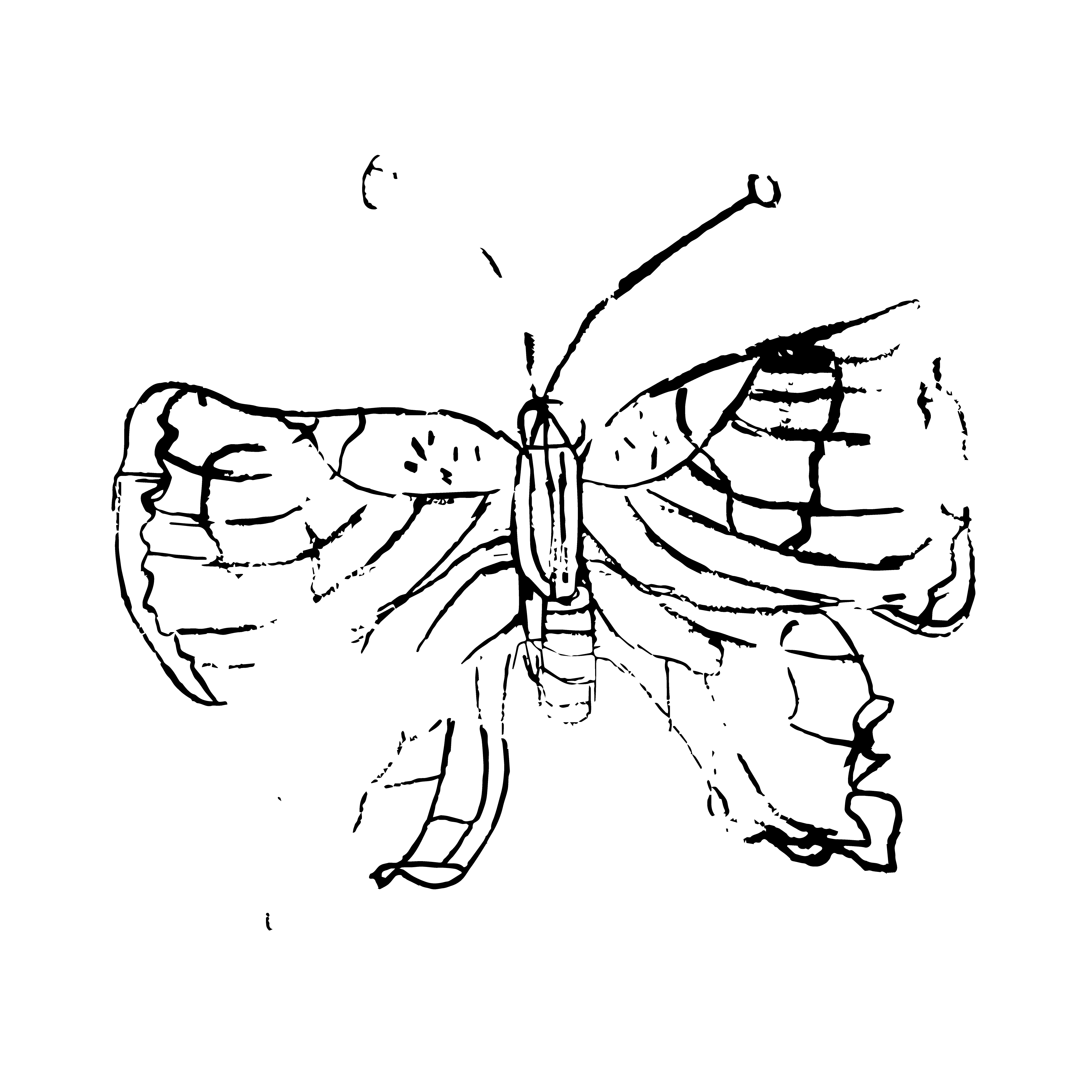}\label{fig:butterfly-adobe}}
\subfloat[][\cite{SimoSerraSIGGRAPH2016}]{\includegraphics[width=0.248\linewidth]{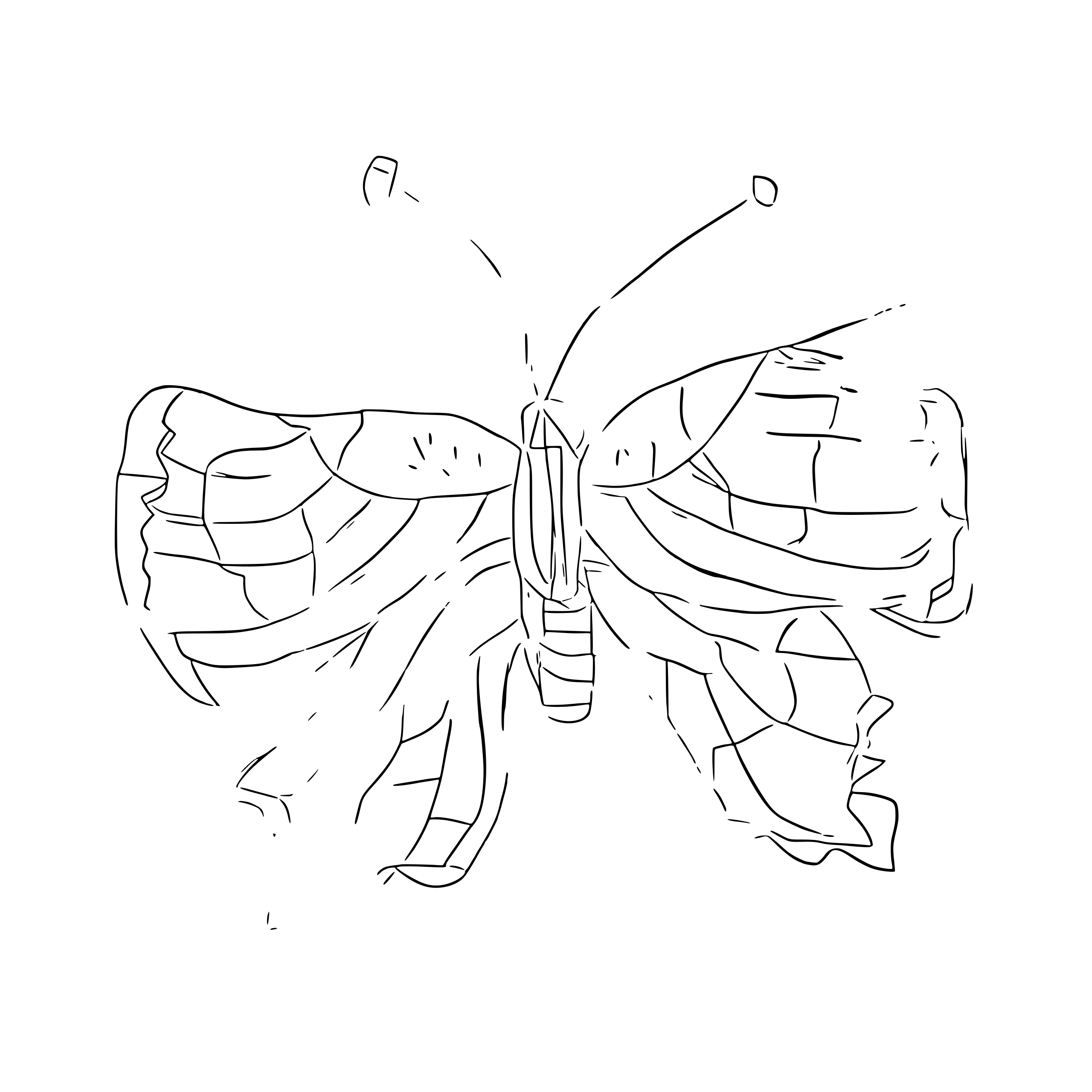}\label{fig:butterfly-simo}}
\subfloat[][Our method]{\includegraphics[width=0.248\linewidth]{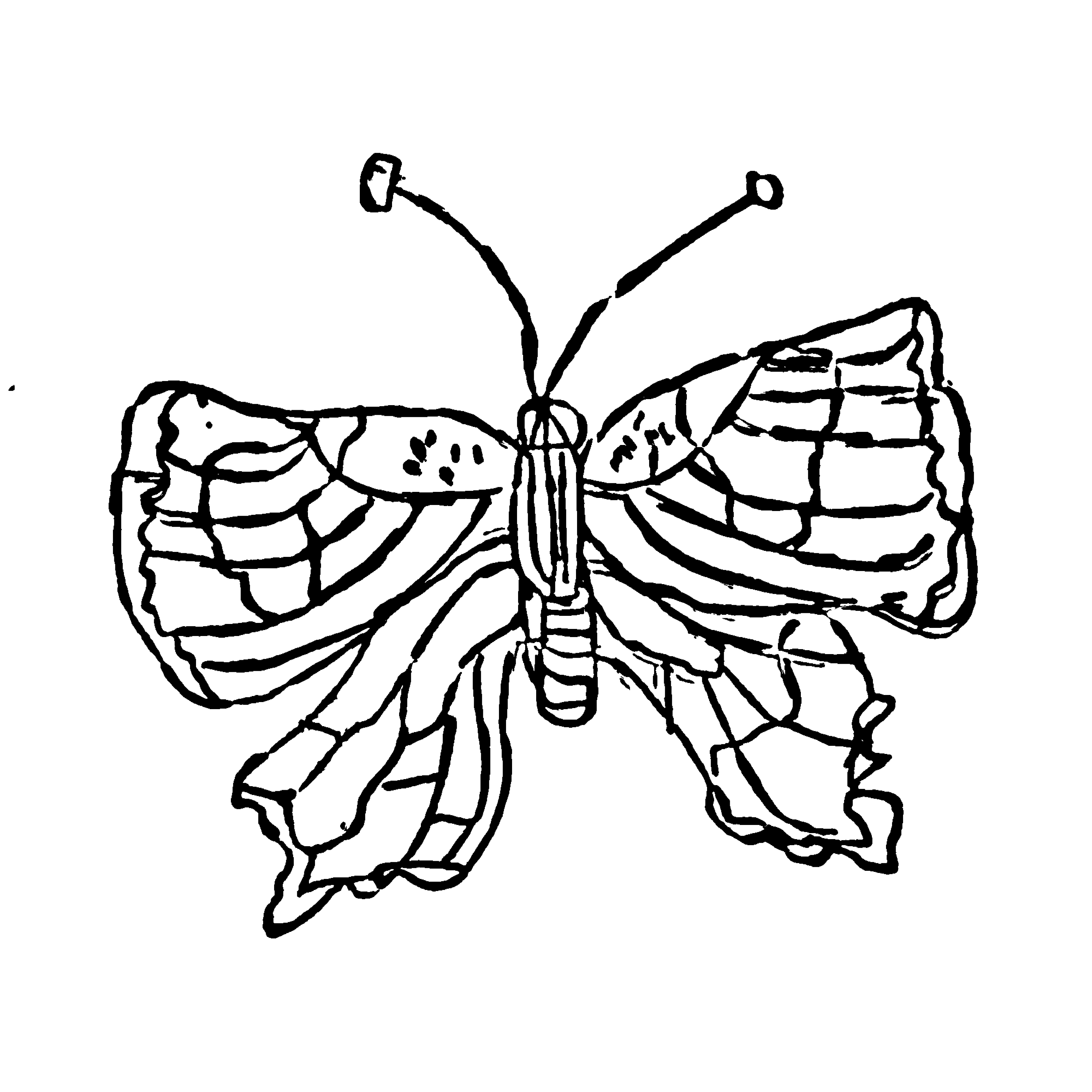}\label{fig:butterfly-our}}

\subfloat[][Input sketch]{\includegraphics[width=0.248\linewidth]{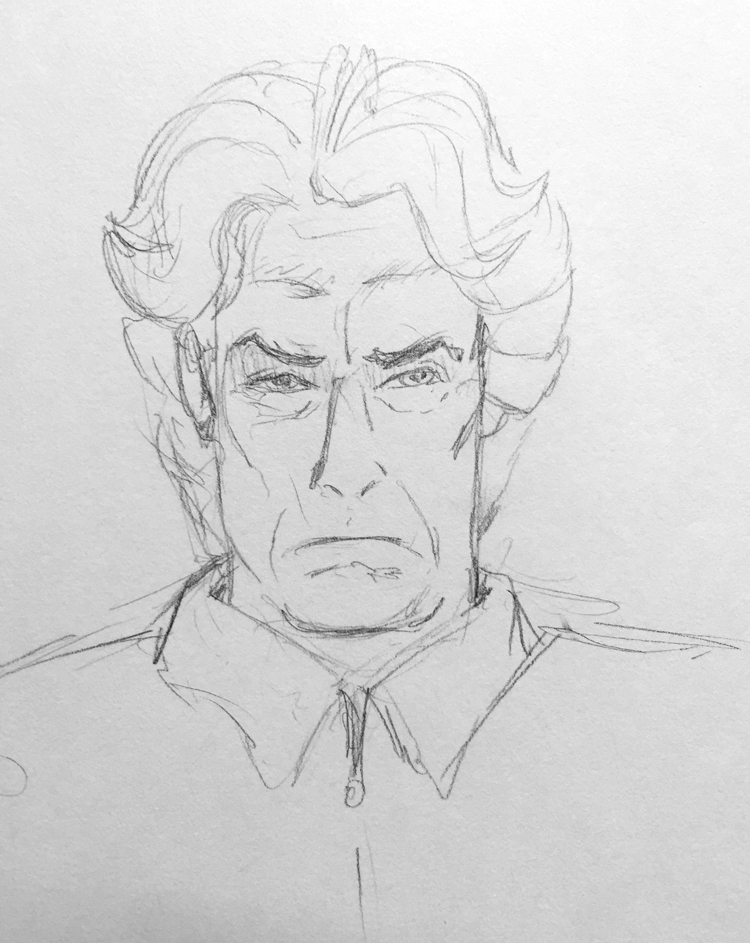} \label{fig:manface}}
\subfloat[][AI\textsuperscript{TM} Live Trace]{\includegraphics[width=0.248\linewidth]{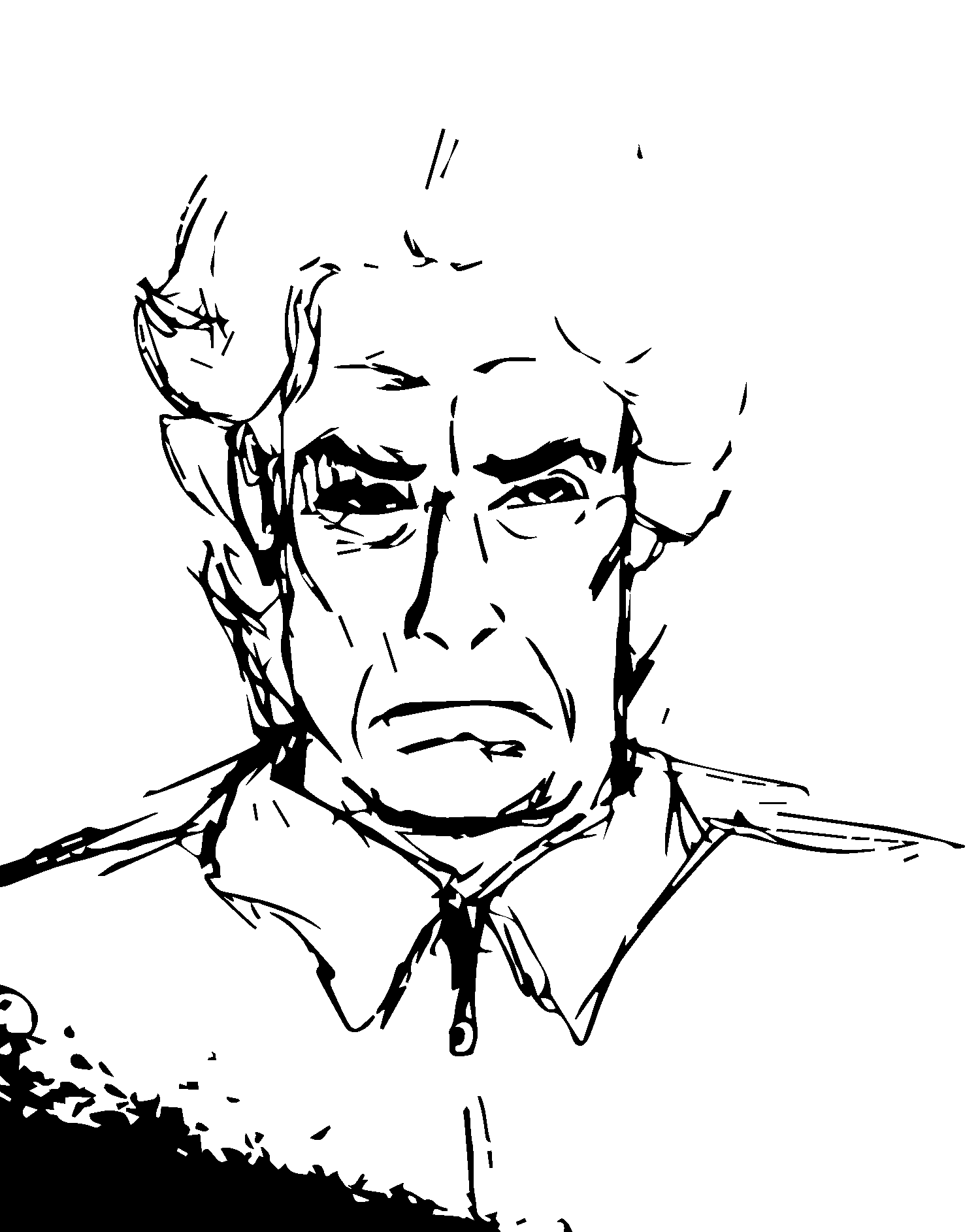}\label{fig:manface-adobe}}
\subfloat[][\cite{SimoSerraSIGGRAPH2016}]{\includegraphics[width=0.248\linewidth]{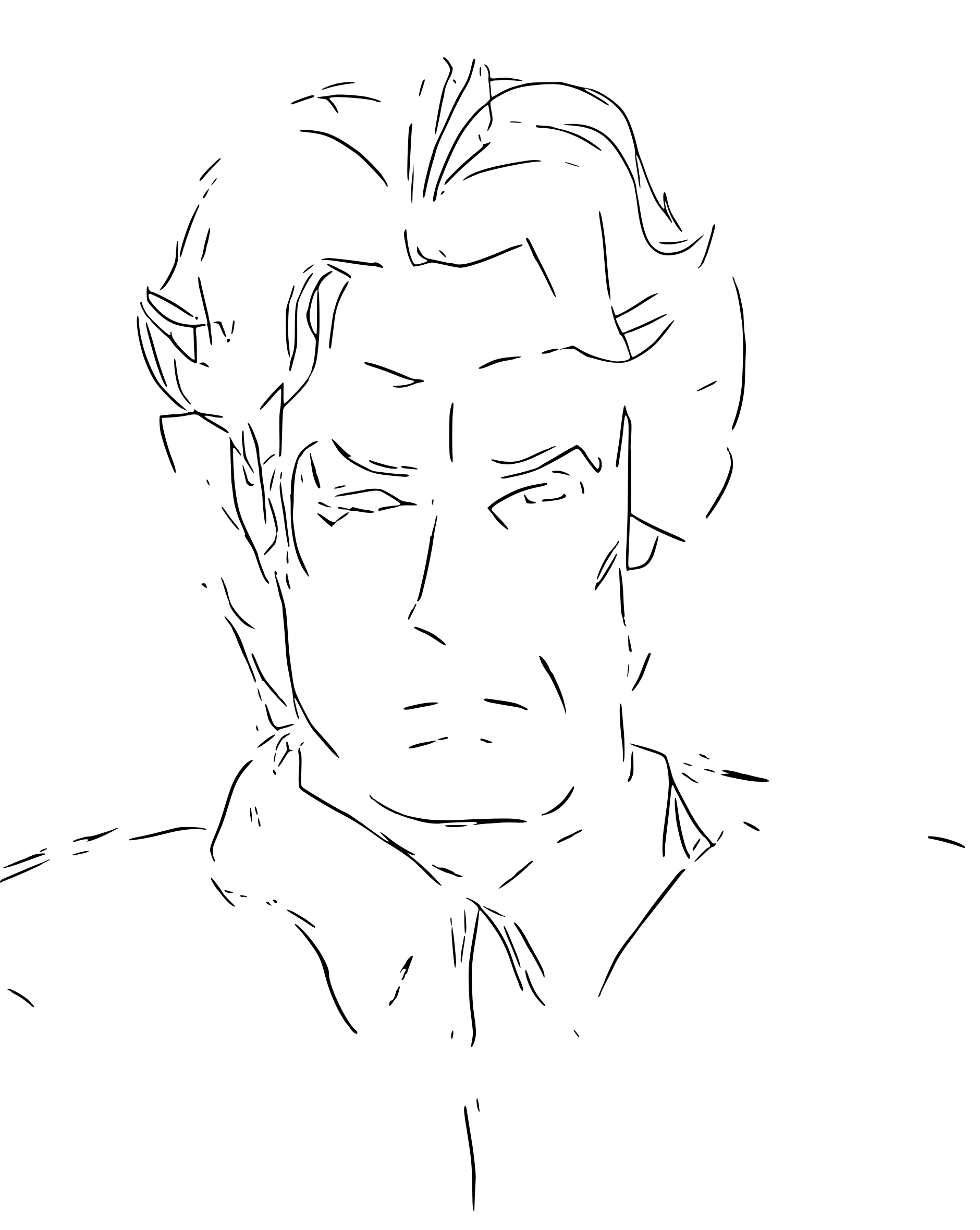}\label{fig:manface-simo}}
\subfloat[][Our method]{\includegraphics[width=0.248\linewidth]{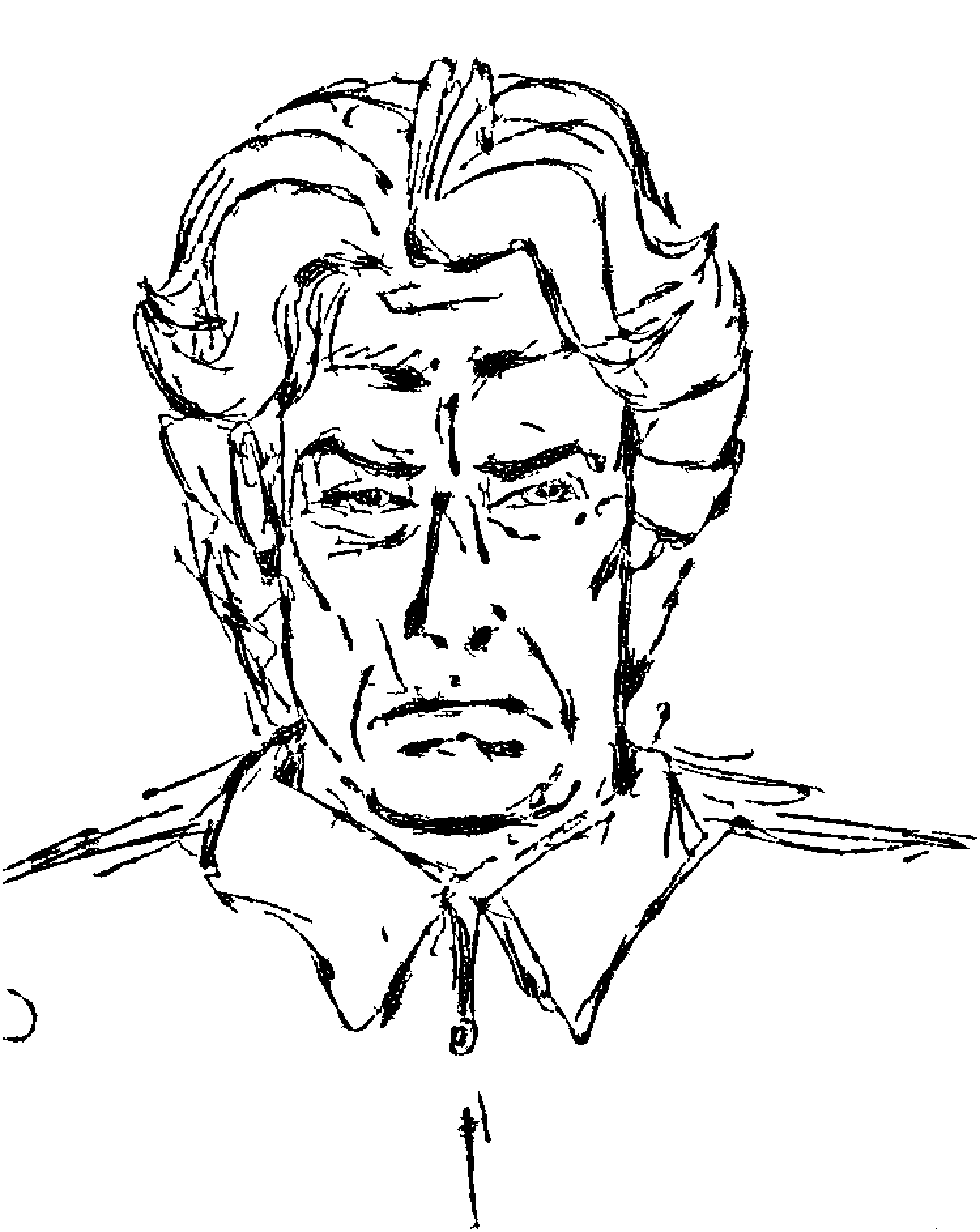}\label{fig:manface-our}}

\subfloat[][Input sketch]{\includegraphics[width=0.248\linewidth]{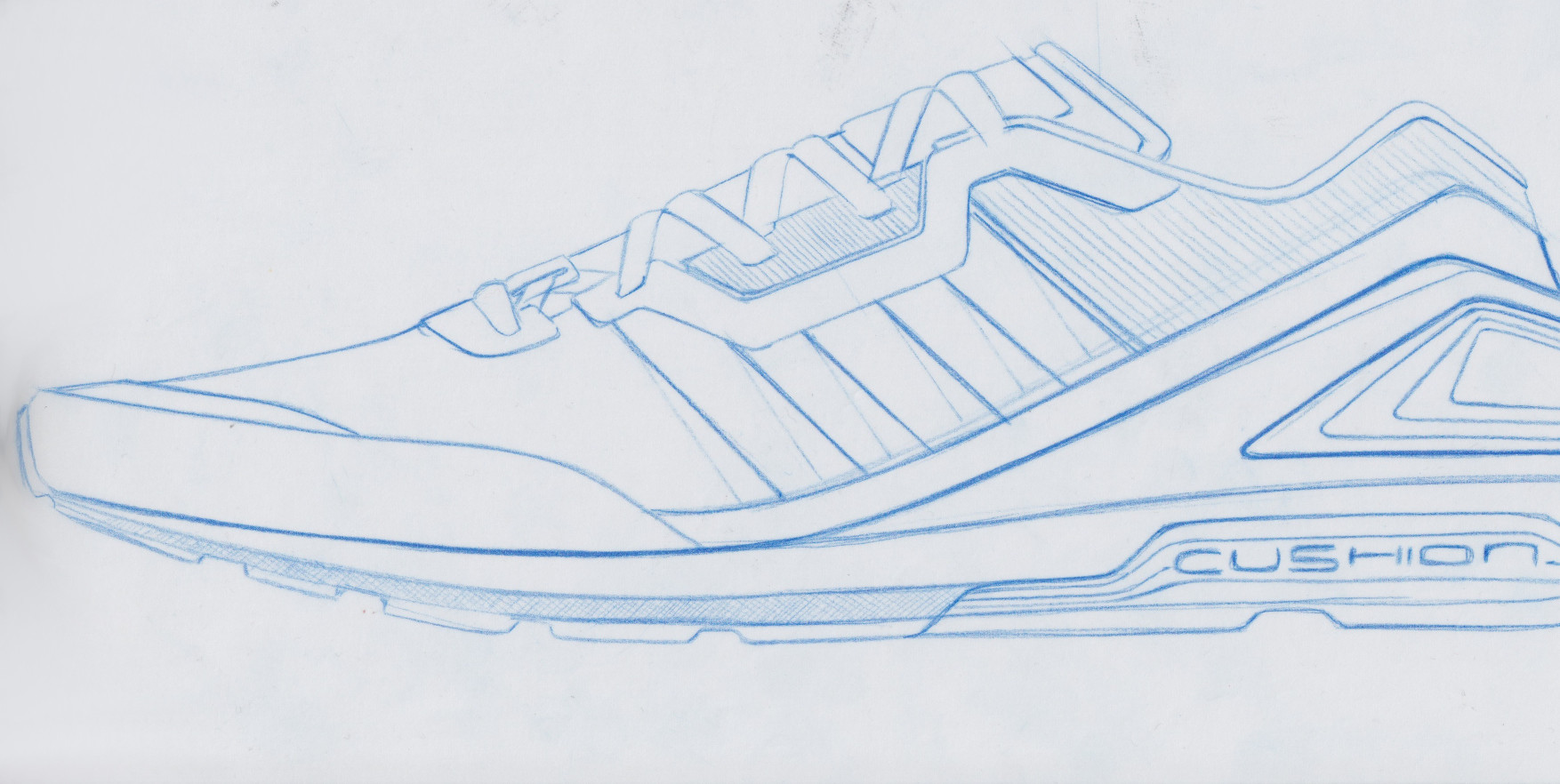} \label{fig:shoe}}
\subfloat[][AI\textsuperscript{TM} Live Trace]{\includegraphics[width=0.248\linewidth]{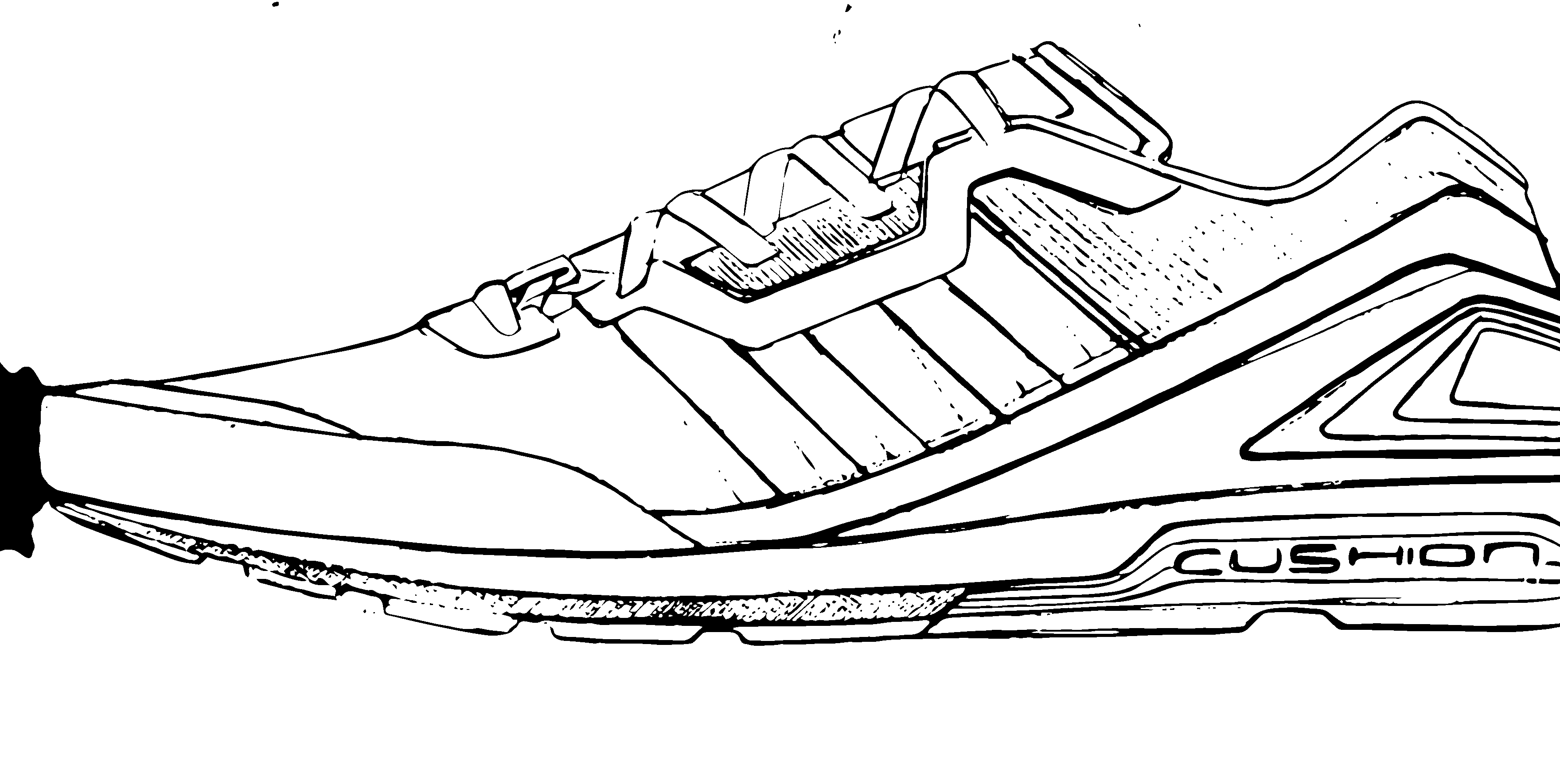}\label{fig:shoe-adobe}}
\subfloat[][\cite{SimoSerraSIGGRAPH2016}]{\includegraphics[width=0.248\linewidth]{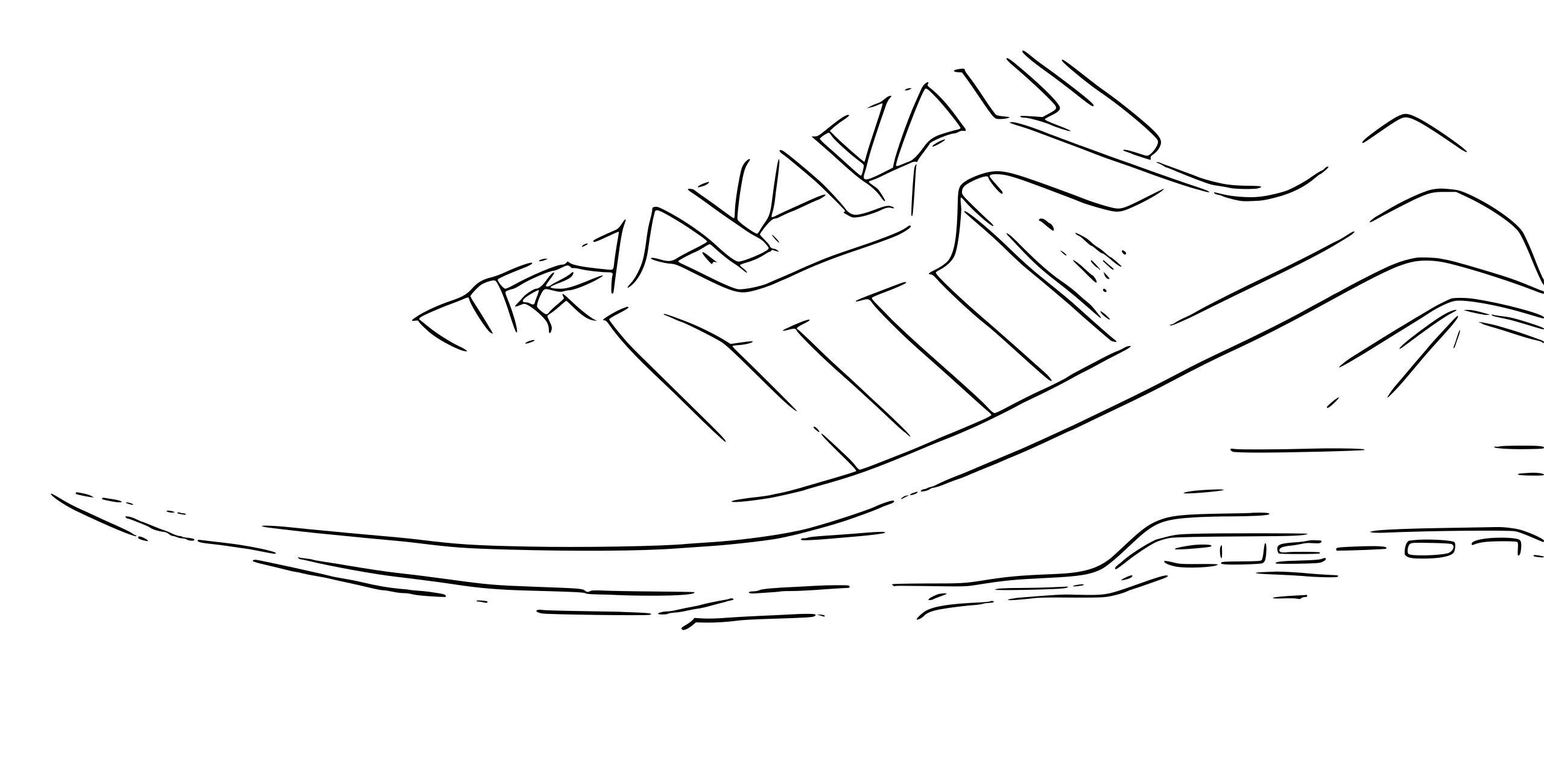}\label{fig:shoe-simo}}
\subfloat[][Our method]{\includegraphics[width=0.248\linewidth]{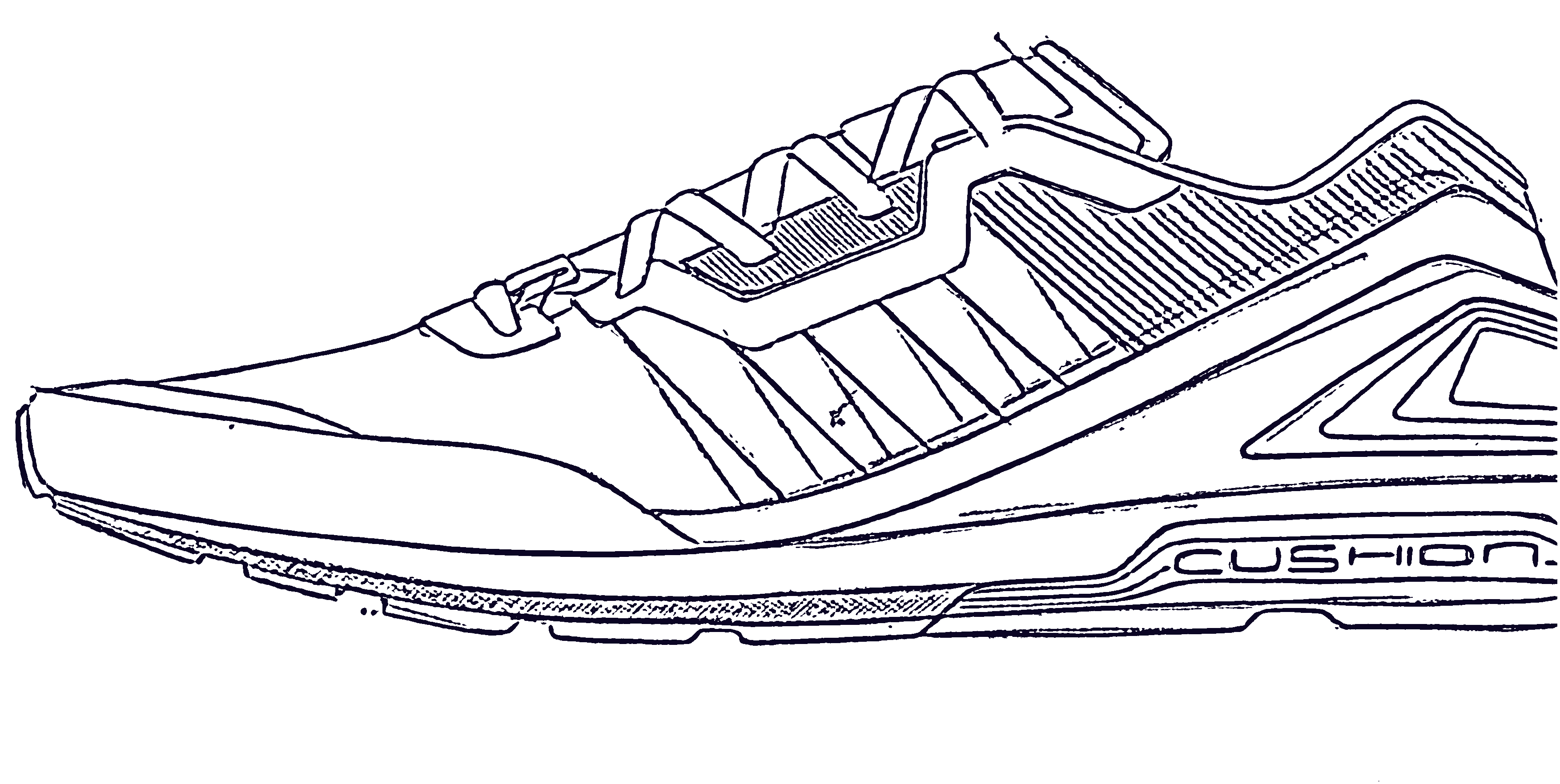}\label{fig:shoe-our}}

\caption{Examples of line extractions from a commercial tool \protect\subref{fig:butterfly-adobe}, \protect\subref{fig:manface-adobe}, \protect\subref{fig:shoe-adobe}, the state of the art algorithm \protect\subref{fig:butterfly-simo}, \protect\subref{fig:manface-simo}, \protect\subref{fig:shoe-simo},  and our method \protect\subref{fig:butterfly-our}, \protect\subref{fig:manface-our}, \protect\subref{fig:shoe-our}. In detail: an image from our inverse dataset \protect\subref{fig:butterfly}; a man's portrait art \protect\subref{fig:manface}, author: Michael Bencik - Creative Commons 4;  a real \financer hand drawn design \protect\subref{fig:shoe}.}
\label{fig:man_butterfly}

\end{figure*}

In addition to that dataset, we also created our own dataset. The motivation relies in the need for a quantitative (together with a qualitative or visual) evaluation of the results. Manually segmenting complex hand-drawn images such as that reported in Fig. \ref{fig:man_butterfly}, last row, to obtain the ground truth to compare with, is not only tedious, but also very prone to subjectivity. With these premises, we searched for large and public datasets (possibly with an available ground truth) to be used in this evaluation. One possible solution is the use of the SHREC13 - ``Testing Sketches'' dataset \cite{li2013shrec}, whereas alternatives are Google Quick Draw and Sketchy dataset \cite{Sangkloy:2016:SDL:2897824.2925954}. SHREC13 contains very clean, single-stroke ``hand-drawn'' sketches (created using a touch pad or mouse), such as those reported in Figs. \ref{fig:bike} and \ref{fig:dogface}. It is a big and representative dataset: it contains 2700 sketches divided in 90 classes and drawn by several different authors. Unfortunately, these images are too clean to really challenge our algorithm, resulting in almost perfect results. The same can be said for Quick Draw and Sketchy datasets. To fairly evaluate the ability of our algorithm to extract lines in more realistic situations, we have created a ``simulated'' dataset, called \emph{inverse dataset}. The original SHREC13 images are used as ground truth and processed with a specifically-created algorithm (not described here) with the aim of ``corrupting'' them and recreating as closely as possible different drawing styles, pencils and paper sheets. More specifically, this algorithm randomly selects portions of each ground truth image, and moves/alters them to generate simulated strokes of different strength, width, length, orientation, as well as multiple superimposed strokes, crossing and broken lines, background and stroke noise. The resulting database has the same size as the original SHREC13 (2700 images), and each picture has been enlarged to 1 MegaPixel to better simulate real world pencil sketches. Example results (used as inputs for our experiments) are reported in Figs. \ref{fig:bike-after} and \ref{fig:dogface-after}.  


We performed visual/qualitative comparisons of our system with the state-of-the-art algorithm reported in \cite{SimoSerraSIGGRAPH2016} and with the Adobe Illustrator\textsuperscript{TM}'s tool ``Live Trace'' using different input images. Regarding the algorithm in \cite{SimoSerraSIGGRAPH2016}, we used their nice online tool where users can upload their own sketches and get the resulting simplified image back for comparison purposes. Results are reported in Fig. \ref{fig:man_butterfly}, where it is rather evident that our method performs better than the compared methods in complex and corrupted areas.

To obtain quantitative evaluations we used the ``inverse dataset''. Precision and recall in line extraction and the \emph{mean Centerline Distance}, similar to the notion of Centerline Error proposed in \cite{noris2013topology}, are used as performance metrics.  
The results are shown in Tab. \ref{table:quality}. Our method outperforms Live Trace and strongly beats \cite{SimoSerraSIGGRAPH2016} in terms of recall, while nearly matching its precision. This can be explained by the focus we put in designing a true image color/contrast-invariant algorithm, also designed to work at multiple resolutions and stroke widths. Simo-Serra \etal low recall performance is probably influenced by the selection of training data (somewhat specific) and the data augmentation they performed with Adobe Illustrator\textsuperscript{TM} (less generalized than ours). This results in a global F-measure of \textbf{97.8\%} for our method w.r.t. \textbf{87.0\%} of \cite{SimoSerraSIGGRAPH2016}. It is worth saying that Simo-Serra \etal algorithm provides better results when applied to clean images, but it shows sub-optimal results applied to real sketches (as shown in Fig. \ref{fig:man_butterfly}, last row).

\begin{table}[b!]
\begin{center}
    \begin{tabular}{ | l | c | c | c |}
    \hline
     & Precision & Recall & Center. Dist.\\ \hline
    Our method & 97.3\% & 98.4\% & 3.58 px\\ \hline
    \cite{SimoSerraSIGGRAPH2016} & 98.6\% & 77.9\% & 4.23 px\\ \hline
    Live Trace & 85.0\% & 83.8\% & 3.73 px\\
    \hline
    \end{tabular}
\caption{Accuracy of the three implementations over the \emph{inverse dataset} generated from SHREC13 \cite{li2013shrec} (2700 images).}
\label{table:quality}
\end{center}
\end{table}

Finally, we compared the running times of these algorithms. We run our algorithm and ``Live Trace'' using an Intel Core i7-6700 @ 3.40Ghz, while the performance of \cite{SimoSerraSIGGRAPH2016} are extracted from the paper. The proposed algorithm is much faster than Simo-Serra \etal method (0.64 sec per image on single thread instead of 19.46 sec on Intel Core i7-5960X @ 3.00Ghz using 8 cores), and offers performance within the same order of magnitude of Adobe Illustrator\textsuperscript{TM} Live Trace (which on average took 0.5 sec). 

\subsection{Unbiased thinning}
\begin{figure*}
\subfloat[Input images]{
\hspace*{-1.5em}
\begin{tabular}[]{c}
  \includegraphics[width=0.245\textwidth]{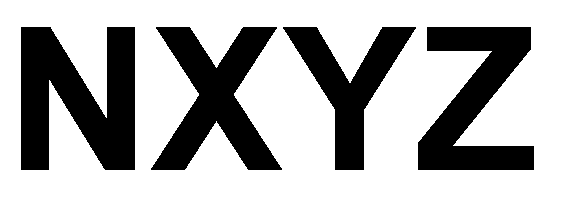}\\
  \includegraphics[width=0.245\textwidth]{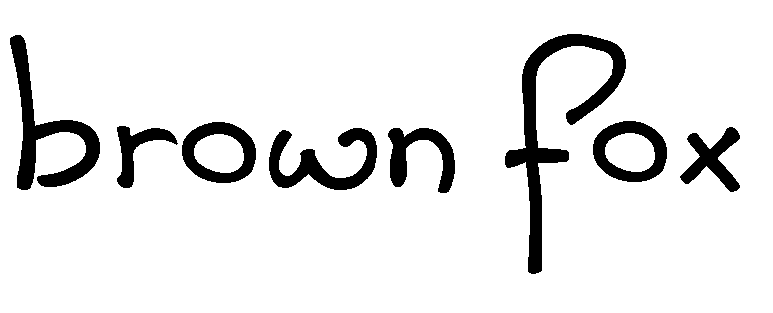}\\
  \includegraphics[width=0.245\textwidth]{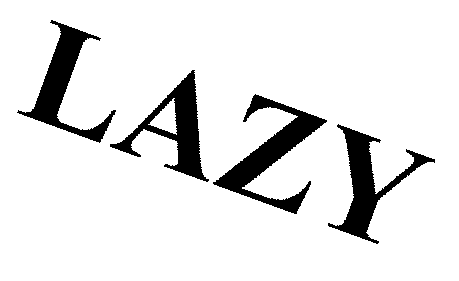}\\
  \includegraphics[width=0.245\textwidth]{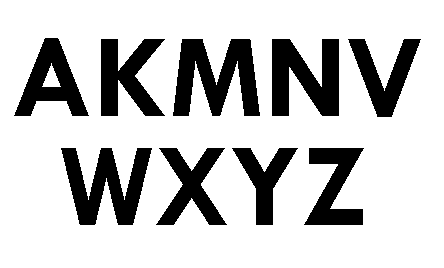}
\end{tabular}
}\hspace*{-1.5em}
\subfloat[\cite{zhang1984fast}]{
\begin{tabular}[]{c}
  \includegraphics[width=0.245\textwidth]{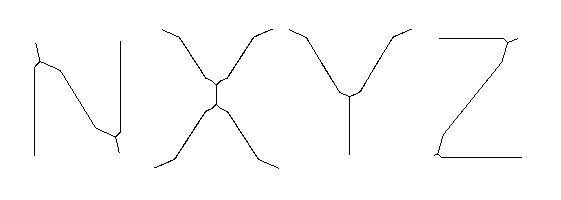}\\
  \includegraphics[width=0.245\textwidth]{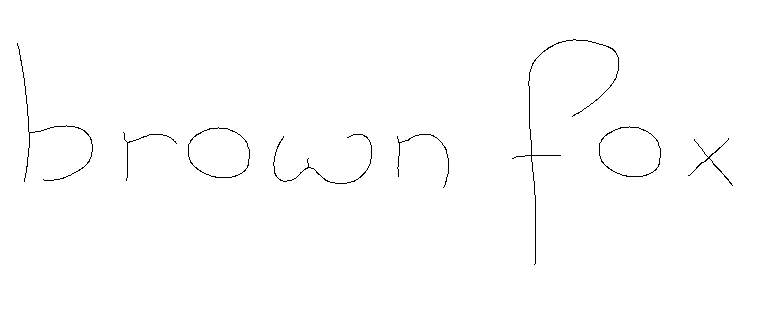}\\
  \includegraphics[width=0.245\textwidth]{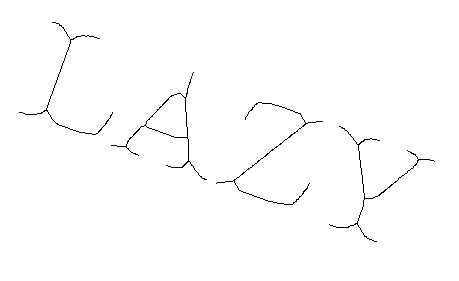}\\
  \includegraphics[width=0.245\textwidth]{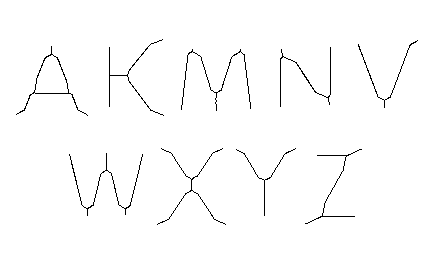}
\end{tabular} 
}\hspace*{-1.5em}
\subfloat[K3M \cite{KhalidSaeed2010}]{
\begin{tabular}[]{c}
  \includegraphics[width=0.245\textwidth]{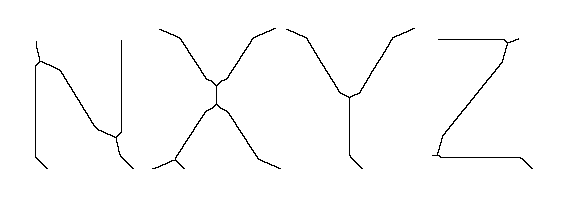}\\
  \includegraphics[width=0.245\textwidth]{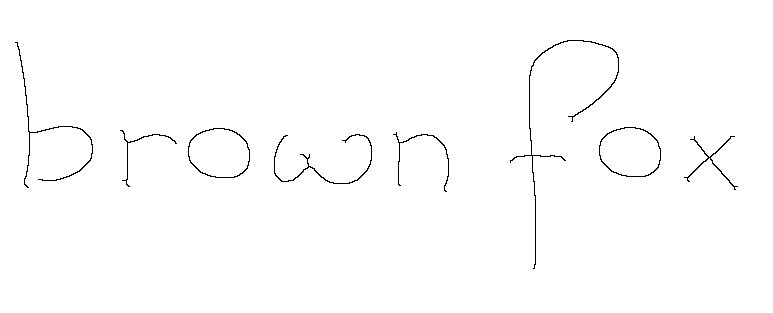}\\
  \includegraphics[width=0.245\textwidth]{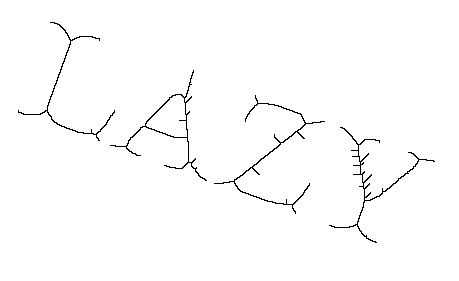}\\
  \includegraphics[width=0.245\textwidth]{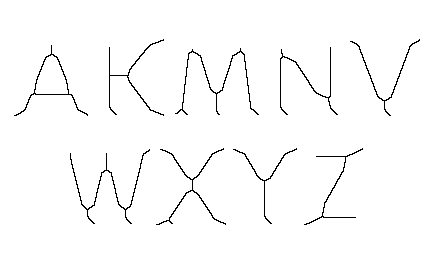}
\end{tabular} 
}\hspace*{-1.5em}
\subfloat[Our]{
\begin{tabular}[]{c}
  \includegraphics[width=0.245\textwidth]{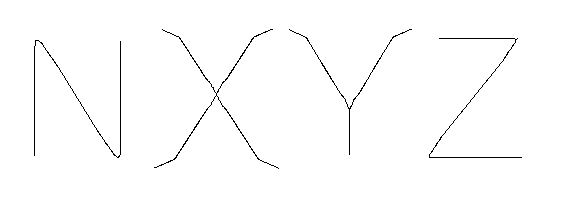}\\
  \includegraphics[width=0.245\textwidth]{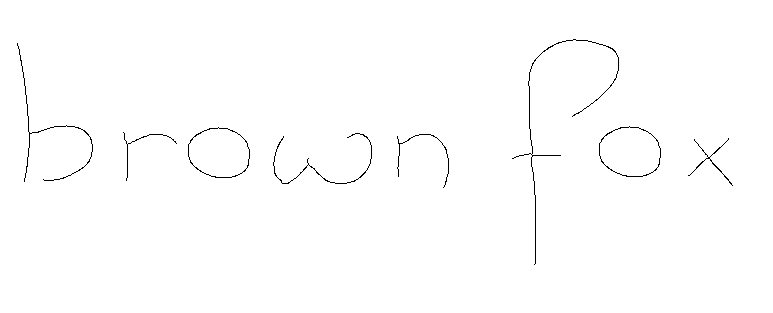}\\
  \includegraphics[width=0.245\textwidth]{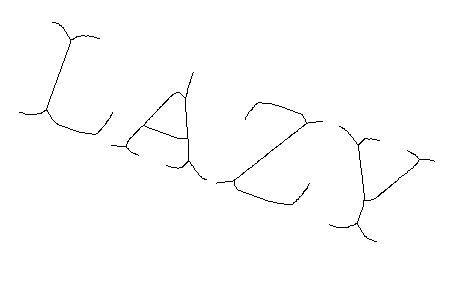}\\
  \includegraphics[width=0.245\textwidth]{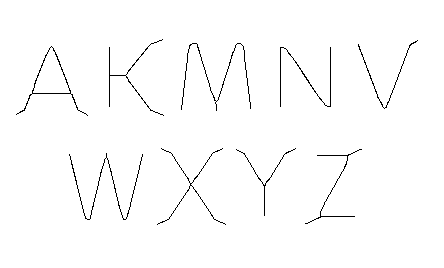}
\end{tabular} 
}
  
  \caption{Examples of thinning results. (a) The input images. (b) Thinning results with the ``standard'' Zhang-Suen algorithm'. (c) Results for K3M method (state of the art). (d) Our algorithm results.
Fonts used (from the top): 200pts Arial; 180pts Dijkstra; 150pts Times New Roman (30 degrees rotated), 120pts TwCen.}
  \label{fig:unbias_results}
\end{figure*}

In order to assess the quality of the proposed unbiased thinning algorithm, a qualitative comparison with other two thinning algorithms is shown in Fig. \ref{fig:unbias_results}. The other two algorithms are: \cite{zhang1984fast}, as a standard, robust, iterative algorithm, and K3M \cite{KhalidSaeed2010}, which represents the state of the art for unbiased thinning. In the literature, the standard testing scenario for thinning algorithms is typewriting fonts. 

It is quite evident that the proposed algorithm is able to correctly thin the most difficult parts of the characters, in particular along ``N'' steep angles, ``X'' crossing, and ``Y'' intersection, where it reconstructs the original structure of the letters with much more precision than the other algorithms.

Moreover, in all of the examples our algorithm associates the robustness of a standard algorithm (like Zhang-Suen), with the complete removal of biases, and has the additional benefit of working with shapes of arbitrary dimension. In fact, the test images range from the dimensions of 120 to 200 font points. K3M, on the other hand, has been designed to reduce bias but does not show any appreciable result with these shapes; this is because it has been designed to work with very small fonts (< 20 pts), and provides limited benefits for larger shapes.

For shapes that are already almost thin, with no steep angles (like Dijkstra cursive font - second row), the thinning results  are similar for all the algorithms.

The unbiasing correction works better when the shapes to be thinned have a constant width (as in  Arial and TwCen fonts). If the width of the line changes strongly within the shape, as in Times New Roman (and in general Serif fonts, that are more elaborated), the unbiasing correction is much more uncertain to perform, and our results are similar to the standard thinning by Zhang-Suen.

We can also note that K3M has some stability problems if the shape is big and has aliased contours, such as the rotated Times New Roman. In that scenario it creates multiple small fake branches.

For every example, our algorithm used a cord length accumulation with size of 15 points.

\subsection{Vectorization algorithm}
We improved Schneider's algorithm mainly regarding the quality of the vectorized result. In particular, our objective was to reduce the total number of control points in the vectorized representation, thus creating shapes that are much easier to modify and interact with. This is sometimes called lines ``weight''.

\begin{figure*}
\begin{center}
\hfill
\subfloat[][Schneider (err = 3): 939 pts]{\includegraphics[width=0.485\textwidth]{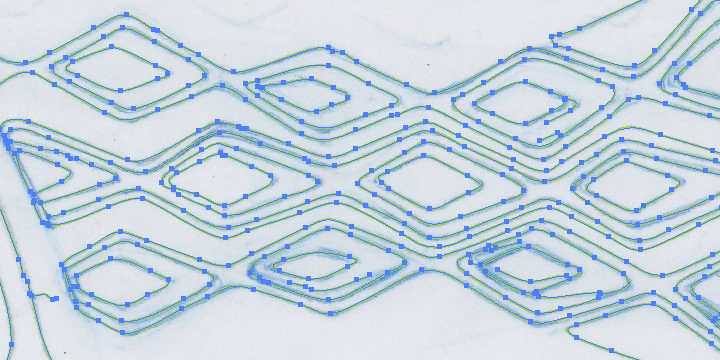}\label{fig:o05}}\hfill
\subfloat[][Our (err = 3): 663 pts]{\includegraphics[width=0.485\textwidth]{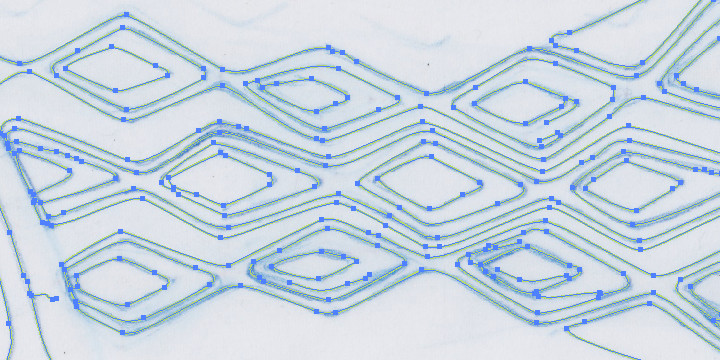}\label{fig:n05}}\hfill
\\
\hfill
\subfloat[][Schneider (err = 6): 590 pts]{\includegraphics[width=0.485\textwidth]{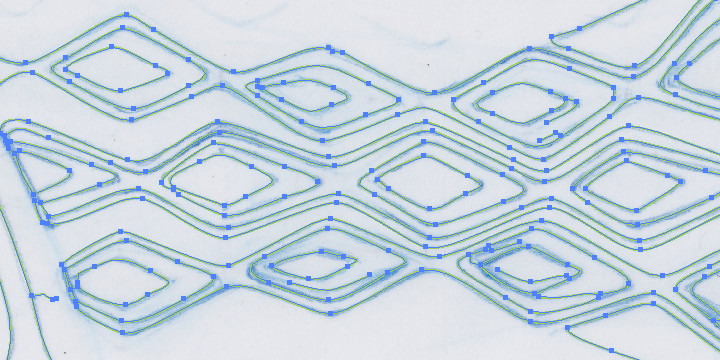}\label{fig:o1}}\hfill
\subfloat[][Our (err = 6): 532 pts]{\includegraphics[width=0.485\textwidth]{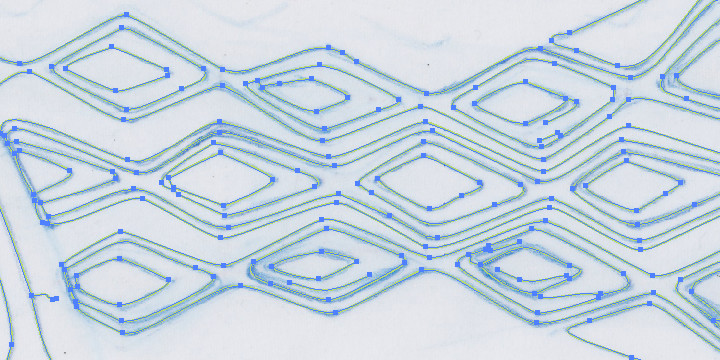}\label{fig:n1}}\hfill
\end{center}
\caption{
Four example portions of vectorization. \protect\subref{fig:o05} and \protect\subref{fig:o1} use Schneider's stock algorithm, \protect\subref{fig:n05} and \protect\subref{fig:n1} use our improved version. Our algorithm considerably reduces the number of control points for both cases (err = 3 and err = 6), without decreasing result quality. For all the examples the maximum number of iterations has been dynamically set for each path to its own length in pixels (1 iteration per pixel).
}
\label{fig:vec_example}
\end{figure*}

Fig. \ref{fig:vec_example} shows that we successfully reduced the number of control points w.r.t. Schneider's original algorithm with a percentage that ranges from 10\% to 30\% less control points, depending on the maximum desired error. This nice property of fewer control points comes at the cost of increased computational time. In fact, the total run time has increased, but is still acceptable for our purposes: standard Schneider's algorithms takes about 1 - 4 seconds to vectorize a 5 MegaPixels sketch on a Intel Core i7-6700 @ 3.40Ghz, whereas the improved version takes 2 - 15 seconds. Moreover, Schneider's algorithm total running time is heavily dependent on the desired error (quality), while our version depends mainly on the specified maximum number of iterations. For our experiments we set the number of iterations dynamically for each path to its own path length, in pixels (short paths are optimized faster than long paths). 

Tabs. \ref{table:err1} and \ref{table:err0_5} show some quantitative analyses about execution times and number of control points obtained. They also show that we could stick with 0.1 iterations per pixel and achieve execution times very similar to the original Schneider's version, while still reducing the number of control points considerably.


\begin{table}[b!]
\begin{center}
    \begin{tabular}{ | l | r | c | c |}
    \hline
              &               & Time (ms) & \# Points \\ \hline
    Schneider & 10   iter/pix & 2860      & 573    \\ \hline
    Schneider &  1   iter/pix & 380       & 590    \\ \hline
    Schneider &  0.1 iter/pix & 85        & 617    \\ \hline
    Our approach      &  1   iter/pix & 14020     & 532    \\ \hline
    Our approach      &  0.1 iter/pix & 1420      & 555    \\ \hline
    \end{tabular}
\caption{Running time and number of control points generated by the two versions of the vectorization algorithm. Desired error $err = 6$. Less is better.}
\label{table:err1}
\end{center}
\end{table}

\begin{table}[b!]
\begin{center}
    \begin{tabular}{ | l | r | c | c |}
    \hline
              &               & Time (ms) & \# Points \\ \hline
    Schneider & 10   iter/pix & 4110      & 904    \\ \hline
    Schneider &  1   iter/pix & 550       & 939    \\ \hline
    Schneider &  0.1 iter/pix & 115       & 983    \\ \hline
    Our approach    &  1   iter/pix & 14735     & 663    \\ \hline
    Our approach      &  0.1 iter/pix & 1483      & 744    \\ \hline
    \end{tabular}
\caption{Running time and number of control points generated by the two versions of the vectorization algorithm. Desired error $err = 3$. Less is better.}
\label{table:err0_5}
\end{center}
\end{table}

\section{Conclusions}\label{sec:conclusions}
The proposed system proved its correctness and viability at treating different input formats for complex hand-drawn sketches. It has been tested with real fashion sketches, artificial generated pictures with added noise, as well as random subject sketches obtained from the web.

The line extraction algorithm outperforms the state of the art in recall, without sacrificing precision.

The unbiased thinning helps in representing shapes more accurately. It has proven to be better than the existing state-of-the-art approaches.

The discussed path extraction algorithm provides a complete treatment of the conversion of thinned images into a more suitable, compact representation.

Finally, Schneider's algorithm for vectorization has been improved in the quality of its results. Experiments show a noticeable reduction in the number of generated control points (by a 10-30\% ratio), while keeping good runtime performance.

In conclusion, the current version of the proposed framework has been made available as an Adobe Illustrator plugin to several designers at Adidas exhibiting excellent results. This further demonstrates its usefulness in real challenging scenarios.

\section*{Acknowledgements}
This work is funded by Adidas AG. We are really thankful to Adidas for this opportunity.

\bibliographystyle{ACM-Reference-Format}
\bibliography{refs}

\end{document}